\documentclass[11pt,a4paper]{article}
\usepackage{a4wide}
\usepackage{graphicx}
\usepackage[fleqn]{amsmath}
\usepackage{amssymb}
\usepackage{amsthm}
\usepackage{bm}
\usepackage{url}
\newcommand{\real}{\mathbb{R}}
\newcommand{\bmf}{\bm{f}}
\newcommand{\bu}{\bm{u}}
\newcommand{\bx}{\bm{x}}
\newcommand{\by}{\bm{y}}

\newcommand{\dd}{\,\mathrm{d}}
\newcommand{\argdot}{\,\cdot\,}
\newcommand{\Div}{\operatorname{div}}
\theoremstyle{plain}

\theoremstyle{remark}

\allowdisplaybreaks
\begin{document}

\title{A Robust Variational Model for Positive Image Deconvolution}
\author{Martin Welk\\
University for Health Sciences,
Medical Informatics and Technology (UMIT),\\
Eduard-Walln\"ofer-Zentrum 1, 6060 Hall/Tyrol, Austria\\
\url{martin.welk@umit.at}
}
\date{October 8, 2013}
\maketitle

\begin{abstract}
In this paper, an iterative method for robust deconvolution 
with positivity constraints is discussed.
It is based on the known variational interpretation of the  
Richardson-Lucy iterative deconvolution as fixed-point iteration for 
the minimisation of an information divergence functional under a
multiplicative perturbation model. 
The asymmetric penaliser function involved in this functional is then
modified into a robust penaliser, and complemented with a regulariser.
The resulting functional gives rise to a fixed point iteration that we call
robust and regularised Richardson-Lucy deconvolution. It achieves an
image restoration quality comparable to state-of-the-art robust variational
deconvolution with a computational efficiency similar to that of the
original Richardson-Lucy method.
Experiments on synthetic and real-world
image data demonstrate the performance of the proposed method.

\bigskip
\noindent%
\textbf{Keywords:}
Non-blind deblurring $\bullet$
Richardson-Lucy deconvolution $\bullet$
Regularization $\bullet$
Robust data term
\end{abstract}

\sloppy
\section{Introduction}

The sharpening of blurred images is a standard problem 
in many imaging applications.
A great variety of different approaches to this severely ill-posed inverse
problem have been developped over time
which differ in the assumptions they make, and in their suitability for 
different application contexts.

\medskip
Blur of an image is described by a \emph{point-spread function (PSF)} 
which describes the redistribution of light energy in the image domain 
$\Omega\subset\real^2$.
When blurring acts equally at all locations, one has a \emph{space-invariant
PSF} $h:\real^2\to\real^+$ which acts by convolution. 
Accounting also for the impact of noise $n$, a typical blur model (with
additive noise) then reads
\begin{equation}\label{genericblurmodelsi}
f = h * g + n\;,
\end{equation}
where $f$ is the observed image, and $g$ the unknown sharp image.

In the more general case of a space-variant blur one needs a point-spread
function with two arguments, 
$H:\varOmega\times\varOmega\to\real^+$, and $h*g$ is replaced with the
integral operator
\begin{equation}
(H \circledast g) (\bx) := \int\limits_\varOmega H(\bx,\by) g(\by) \dd\by
\label{intopH}
\end{equation}
such that 
\begin{equation}\label{genericblurmodel}
f = H \circledast g + n \;,
\end{equation}
which subsumes the space-invariant case by setting 
$H(\bx,\by)=h(\bx-\by)$.
We denote by $H^*$ the adjoint of the point-spread function $H$,
which is given by $H^*(\bx,\by):=H(\by,\bx)$.
Conservation of energy implies generally that $H^*\circledast1=1$,
however, this condition may be violated near image boundaries due
to blurring across the boundary.

In deblurring,
we want to obtain a restored image $u$ that approximates $g$, with
the degraded image $f$ and the PSF $H$ as input.
This is the case of \emph{non-blind} deconvolution (as opposed to blind
deconvolution which aims at inferring the sharp image and the
PSF simultaneously from the degraded image).

Some approaches to the deconvolution problem are presented in more
detail in Section~\ref{sec-exdcvm}.

\paragraph{Our contribution.}
The main subject of this paper is to discuss a modification of the
Richardson-Lucy (RL) deconvolution method \cite{Lucy-AJ74,Richardson-JOSA72} 
by robust data terms. Building on the known variational interpretation of RL 
deconvolution \cite{Snyder-TIP92}, we replace the asymmetric penaliser
function in the data term in such a way that larger residual errors are
penalised less than with the standard Csisz\'ar divergence term.

Using robust data terms together with a regulariser similar to 
\cite{Dey-isbi04,Dey-MRT06} we obtain a robust and regularised Richardson-Lucy
variant that unites the high restoration quality of variational
deconvolution methods with high efficiency that is not far from the
original Richardson-Lucy iteration.
This method has already been used for experiments on recovering information
from diffuse reflections in \cite{Backes-sp09} and, combined
with interpolation, for the enhancement of confocal microscopy images
\cite{Elhayek-dagm11,Persch-MST13x}. It has also been used to achieve
efficient deconvolution under real-time or almost-real-time conditions
\cite{Welk-aapr13,Welk-SIVP13}.
We demonstrate that both robust data terms and regularisers contribute
substantially to its performance.

An earlier version of the present work is the technical report 
\cite{Welk-tr10}.

\paragraph{Related work.}
The omnipresence of deblurring problems has made 
researchers address this problem since long
\cite{vanCittert-ZPhys33,Wiener-Book49}.
From the abundant literature on this topic, the most relevant work
in our present context includes Richardson-Lucy deconvolution
\cite{Lucy-AJ74,Richardson-JOSA72}, variational methods 
\cite{Chan-TIP98,Osher-icip94,Rudin-PHYSD92,You-icip96}, and their interplay 
\cite{Dey-isbi04,Dey-MRT06,Snyder-TIP92}; see also \cite{Bratsolis-AA01} 
for another approach to combine Richardson-Lucy deconvolution with 
regularisation.

Fundamental theoretical results on existence and uniqueness of solutions of
deconvolution problems can be found in the work of Bertero et al.\ 
\cite{Bertero-PIEEE88}.

Robust data terms in deconvolution go back to Zervakis et al.\ 
\cite{Zervakis-TIP95} in statistical models, and have recently been used 
intensively 
in the variational context by Bar et al.\ \cite{Bar-scs05,Bar-ssvm07} and
Welk et al.\ \cite{Welk-ibpria07,Welk-dagm05}. Positivity 
constraints were studied in discrete iterative deconvolution by Nagy and 
Strako{\v s} \cite{Nagy-spie00} and in a variational framework by Welk and 
Nagy \cite{Welk-ibpria07}. The extension of variational approaches to 
multi-channel images has been studied in \cite{Gerig-TMI92} and more 
specifically in deconvolution in \cite{Bar-vlsm05,Welk-ibpria07}.

\paragraph{Structure of the paper.}
In Section~\ref{sec-exdcvm} we recall approaches to
image deconvolution which form the background for our approach, and
discuss some aspects of noise models and robustness. 
Section~\ref{sec-rrrl} recalls the embedding of Richardson-Lucy deconvolution
into a variational context. Exploiting this connection, the RL algorithm can be
modified in order to increase restoration quality and robustness with respect
to noise and perturbations. An experimental comparison of the deconvolution
techniques under consideration is provided in Section~\ref{sec-exp} based on
both synthetic and real-world data. Conclusions in Section~\ref{sec-conc}
end the paper.

\section{Deconvolution Methods}\label{sec-exdcvm}

We start by recalling selected deconvolution approaches from the
literature which we will refer to later.

\subsection{Richardson-Lucy Deconvolution}

Richardson-Lucy (RL) deconvolution 
\cite{Lucy-AJ74,Richardson-JOSA72} is a nonlinear iterative method
originally motivated from statistical considerations. It is based
on the assumption of positive grey-values and Poisson noise
distribution.
If the degraded and sharp images, and the point-spread function 
are smooth functions over $\real^2$ with positive real 
values, one uses the iteration
\begin{equation}
\label{RL}
u^{k+1} = \left(H^* \circledast \frac{f}{H \circledast u^k}
\right)\cdot u^k
\end{equation}
to generate a sequence of successively sharpened images $u^1,u^2,\ldots$
from the initial image $u^0:=f$. 

In the absence of noise the sharp image $g$ is a fixed
point of \eqref{RL}, as in this case the multiplier 
$H^*\circledast \frac{f}{H\circledast g}$
equals the constant function $1$. 

The single parameter of the procedure is the 
number of iterations. While with increasing number of iterations greater 
sharpness is achieved, the degree of regularisation is reduced,
which leads to amplification of artifacts that in the long run
dominate the filtered image. This phenomenon is known by the
name of \emph{semi-convergence}.

\subsection{Variational Deconvolution}

Variational methods \cite{Osher-icip94,You-icip96} address the 
deconvolution task by 
minimising a functional that consists of two parts: a \emph{data term}
that enforces the match between the sought image and the observed 
image via the blur model, and a \emph{smoothness term} or \emph{regulariser}
that brings in regularity assumptions about the unknown sharp image. 
The strength of variational approaches lies in their great flexibility, and
in the explicit way of expressing the assumptions made. They achieve
often an excellent reconstruction quality, but their computational
cost tends to be rather high.

A general model for variational deconvolution of a grey-value image
$u$ with known point-spread function $H$ is based on minimising
the energy functional
\begin{equation}\label{VDCVE}
E_{f,H}[u] = \int\limits_\varOmega \left(
\varPhi\bigl((f-H\circledast u)^2\bigr) +
\alpha\, \varPsi\bigl( \lvert\nabla u\rvert^2 \bigr) \right) \dd \bx
\end{equation}
in which the data term $\varPhi\bigl((f-H\circledast u)^2\bigr)$ 
penalises the reconstruction error or \emph{residual} $f-H\circledast u$ 
to suppress deviations from the blur model, while the smoothness term
$\varPsi\bigl(\lvert\nabla u\rvert^2\bigr)$ penalises roughness of the 
reconstructed image. The \emph{regularisation weight} $\alpha$ balances the 
influences of both terms.
$\varPhi, \varPsi:\real^+_0\to\real^+_0$ are increasing penalty functions.

In the simplest case, $\varPhi$ is the identity, thus imposing a quadratic
penalty on the data model. Doing the same in the regulariser, one has
Whittaker-Tikhonov regularisation \cite{Tikhonov-SMD63,Whittaker-PEMS23} 
with $\varPsi(s^2)=s^2$. As this choice leads to a blurring that directly 
counteracts the desired sharpening, it is often avoided in favour of
regularisers with edge-preserving properties. 
A popular representative is total variation deconvolution 
\cite{Chan-TIP98,Osher-icip94,Vogel-TIP98} which uses
$\varPsi(s^2)=\lvert s\rvert$. 
Edge-enhancing regularisers (e.g.\ of the Perona-Malik \cite{Perona-PAMI90}
type) were studied in \cite{Bar-eccv04,Welk-TBW-scs05}.
In \cite{Jung-TIP11,Jung-JCAM13,Jung-ssvm09}, nonlocal regularisation was
proposed.

For the actual minimisation, often 
gradient descent or lagged-diffusivity-type minimisation schemes are 
employed. Iterative schemes with advantageous convergence behaviour
are derived from a half-quadratic approach in \cite{Wang-SIIMS08} 
(for the total variation regulariser) and \cite{Krishnan-nips09}
(for certain non-convex regularisers).

Using in \eqref{VDCVE} a penaliser $\varPhi$ with less-than-quadratic
growth leads to \emph{robust data terms} 
\cite{Bar-scs05,Zervakis-TIP95}. For example,
\cite{Bar-scs05} use the $L^1$ penaliser $\varPhi(s)=\lvert s\rvert$.
The concept of robustness originates from statistics 
\cite{Huber-book81}, and will be discussed in more detail below.
Spatially variant robust deconvolution models were investigated in
\cite{Bar-ssvm07,Welk-dagm05}.

Even Richardson-Lucy deconvolution 
can be interpreted as a fixed point iteration for an optimisation problem
\cite{Snyder-TIP92}, using
Csisz{\'a}r's \emph{information divergence} \cite{Csiszar-AS91}
as an (asymmetric) penaliser function. In a space-continuous
formulation, one has the energy functional
\begin{equation}
\label{RLfunc}
E_{f,H}[u] := \int\limits_{\varOmega} 
\left(H\circledast u-f-f\ln\frac{H\circledast u}{f}\right)
\dd\bx \;.
\end{equation}
No regularisation term is included, which is linked to
the above-mentioned semi-convergence behaviour.
Nevertheless, the energy formulation permits to modify 
Richardson-Lucy deconvolution in the same flexible manner as the 
standard variational approach \eqref{VDCVE} by introducing robust 
data terms and edge-preserving, or even edge-enhancing, regularisers.
While an edge-preserving regulariser in combination with 
Richardson-Lucy deconvolution has been used by Dey et al.\ 
\cite{Dey-isbi04,Dey-MRT06} (in a space-invariant setting), the
possibility of a robust data term has so far not been studied in 
detail, although it has been used successfully in applications 
\cite{Backes-sp09,Elhayek-dagm11,Persch-MST13x,Welk-aapr13,Welk-SIVP13}.

\subsection{Data Terms, Noise Models and Robustness}

Statistical considerations \cite{Geman-PAMI84,Zervakis-TIP95} 
link particular data terms to specific noise models,
in the sense that minimising the so constructed
energy functional yields a maximum likelihood estimator under 
the corresponding type of noise.
For example, quadratic penalisation $\varPhi(s^2)=s^2$ pertains
to Gaussian noise, while the $L^1$ penaliser 
$\varPhi(s^2)=\lvert s\rvert$ matches Laplacian noise. The 
asymmetric penaliser in \eqref{RLfunc} is related to Poisson noise.

This relation allows to use an optimally adapted energy model 
whenever one has full control over the imaging process and can 
therefore establish an accurate noise model.
In practice, one does not always have
perfect control and knowledge of the imaging process.
This is where the concept of \emph{robustness} comes into play.
According to Huber \cite[p.~1]{Huber-book81}, ``robustness signifies 
insensitivity to small deviations from the assumptions''. In 
particular, ``distributional robustness'' means that ``the shape of 
the underlying distribution deviates slightly from the assumed 
model'' \cite[p.~1]{Huber-book81}.
This obviously applies to imaging processes in which
no exact noise model is known, but also further violations of model 
assumptions can be subsumed here, such as imprecise estimates of PSF, 
or errors near the image boundary due to blurring across the 
boundary, see \cite{Welk-dagm05}. 

To incorporate each single influence factor into a model is not
always feasible. Robust models are designed to cope with remaining 
deviations, and still produce usable results. In view of the uncertainty 
about the true distribution of noise, they are often 
based on data terms that match types of noise that are assumed to 
be ``worse'' than the real noise.
A crucial point is to suppress the effect of outliers, which is 
achieved e.g.\ by data terms that penalise outliers less. 
Models are then adapted to distributions with ``pessimistically'' 
heavy tails. 

To evaluate robustness experimentally, it is not only legitimate but even
necessary to test such models against severe, maybe even unrealistic,
types of noise that do not exactly match the model. A frequently used
test case is impulse noise, and it turns out that
variational approaches actually designed for Laplacian noise can 
cope with it practically well \cite{Bar-scs05}.
Of course, one can no longer expect to establish optimality in the 
maximum-likelihood sense with respect to the true noise.

Furthermore, any comparison between methods that are optimised for
different noise models inevitably involves testing at least one of
them with non-matching noise. For example, this happens already in any
comparison of Wiener filtering (Gaussian noise) against Richardson-Lucy 
(Poisson noise).

\subsection{Gradient Descent Methods}

Functionals of type \eqref{VDCVE} are often minimised using gradient descent
\cite{Welk-TBW-scs05,Welk-dagm05}. Alternatively, elliptic iteration schemes
can be used \cite{Bar-scs05}. 

\paragraph{Standard gradient descent.}
To derive a gradient descent equation for the energy $E:=E_{f,H}$ from
\eqref{VDCVE}, one computes by the usual Euler-Lagrange formalism
$\delta_v E:=\left.\frac{\mathrm{d}}{\mathrm{d}\varepsilon}
E[u+\varepsilon v]\right|_{\varepsilon=0}$ for a small additive 
perturbation $v$. 
The variational gradient (G{\^{a}}teaux derivative) $\delta E/\delta u$ is
obtained from $\delta_v E$ by the requirement that
$\left\langle \frac{\delta E}{\delta u},v\right\rangle=\delta_v E$ 
holds for all test functions $v$, where $\langle\,\cdot\,,\,\cdot\,\rangle$
is the standard inner product of functions.

In an elliptic approach, one now solves the Euler-Lagrange
equation $\delta E/\delta u=0$. A gradient descent for \eqref{VDCVE}
is given by the integro-differential equation (compare \cite{Welk-dagm05})
\begin{align}\label{VGD}
\partial_t u = - \frac{\delta E}{\delta u}
&= H^* \circledast \Bigl(
\varPhi'\bigl((f-H\circledast u)^2\bigr) \cdot (f-H\circledast u)\Bigr) 
+
\alpha \Div \left( \varPsi'\bigl(\lvert\nabla u\rvert^2\bigr)\nabla u\right)
\;.
\end{align}
Complementing this equation with e.g.\ the blurred image $f$ as initial
condition and suitable boundary conditions,
one has an initial-boundary value problem which is then approximated
numerically until a steady state is reached, for example via an explicit 
Euler scheme. As each iteration involves two $\circledast$ operations,
the computational expense is fairly high. Even more sophisticated approaches 
like semi-implicit schemes with conjugate gradients for \eqref{VGD}
or the elliptic approach do not change the computational cost substantially,
as the number of iterations for convergence in any scheme depends primarily 
on the extent and structure of the PSF.

\paragraph{Positivity-constrained gradient descent.}
From modelling considerations one can often infer additional information
that helps to mitigate the ill-posedness of the deconvolution problem.
For instance, in a scalar-valued image with grey-values proportional to
radiance, only positive values are admissible.

A strategy to incorporate this positivity constraint in the minimisation of
\eqref{VDCVE} has been described in \cite{Welk-ibpria07}. Based on earlier 
work \cite{Nagy-spie00}, it is proposed to reparametrise the grey-values
by the substitution $u=\exp(z)$. The values of the new image function $z$
are unconstrained in $\real$. Instead of \eqref{VGD} one obtains
thereby the gradient descent
\begin{equation}\label{VGDphi}
\partial_t u = - u \cdot \frac{\delta E}{\delta u}
\;.
\end{equation}

As detailed in \cite{Welk-ibpria07}, 
performing the gradient descent based on the variation of
the function $z$ comes down to a function space transformation. 
Essentially, the positivity-constrained gradient
descent turns out to be the gradient descent in a hyperbolic metric.
From this viewpoint, zero and negative grey-values are avoided because
they are put at infinite distance from any positive values.
This reinterpretation can immediately be transferred to interval constraints
with a suitable metric on the open interval $(a,b)$.

We want to point out here another reinterpretation which is obtained by using
a \emph{multiplicative} gradient descent. To this end, one
considers a multiplicative perturbation with a test function $v$ and
computes the variation
$\delta_v^*E:=
\left.\frac{\mathrm{d}}{\mathrm{d}\varepsilon}E[u(1+\varepsilon v)]
\right|_{\varepsilon=0}$. Analogous to the ordinary G{\^{a}}teaux derivative
$\delta E/\delta u$ above, a ``multiplicative gradient'' $G$ is then
derived from the requirement $\delta_v^*E=\left\langle G,v\right\rangle$.
One finds $G=u\cdot \delta E/\delta u$, which constitutes a
new derivation of the gradient descent \eqref{VGDphi} 
without the substitution of $u$.

\section{Robust and Regularised Richardson-Lucy Deconvolution}\label{sec-rrrl}

For given $f$, $H$, equation \eqref{RL} 
can be understood as a fixed point iteration associated to the minimisation 
of the functional
\eqref{RLfunc},
compare \cite{Snyder-TIP92}. 
This is the so-called \emph{information divergence} introduced by
Csisz\'ar \cite{Csiszar-AS91}.
The asymmetric penaliser function 
\begin{equation}
r_f(w)=w-f-f \ln \frac wf
\end{equation}
is strictly convex for $w>0$ with its minimum at $w=f$.

As a necessary condition for $u$ to be a minimiser of \eqref{RLfunc},
one can compute an Euler-Lagrange equation which in this case becomes 
particularly simple as no derivatives of $u$ are present in the integrand. 
In view of the positivity requirement for $u$ we start by a
multiplicative perturbation of \eqref{RLfunc} with a test function $v$,
\begin{align}
\left.\frac{\mathrm{d}}{\mathrm{d}\varepsilon}E_{f,H}[u(1+\varepsilon v)]
\right|_{\varepsilon=0} 
&
=
\frac{\mathrm{d}}{\mathrm{d}\varepsilon} 
\int\limits_{\varOmega} 
\biggl(H\circledast \bigl(u(1+\varepsilon v)\bigr)
-f-
f\ln\frac{H\circledast \bigl(u(1+\varepsilon v)\bigr)}{f}\biggr)
\dd\bx
\bigg|_{\varepsilon=0}
\notag
\\
&=
\int\limits_\Omega \left(1-\frac{f}{H\circledast u}\right)\,
\bigl(H\circledast (uv)\bigr)
\dd\bx
\;.
\end{align}
With \eqref{intopH} for the integral operator $\circledast$ this becomes
\begin{equation}
\int\limits_\varOmega \int\limits_\varOmega 
\left(1-\frac{f}{H\circledast u}\right)(\bx)\,
H(\bx,\by)\,u(\by)v(\by) \dd\by \dd\bx
\end{equation}
and, after changing the order of integration and rewriting $H$ into $H^*$,
\begin{align}
\int\limits_\varOmega \int\limits_\varOmega 
\left(1-\frac{f}{H\circledast u}\right)(\bx)\,
H^*(\by,\bx)\,u(\by)v(\by) \dd\bx \dd\by
&
=
\int\limits_\varOmega
\left(H^* \circledast \left(1-\frac{f}{H\circledast u}\right)
\right)uv \dd\bx \;.
\end{align}
Requiring that this expression vanishes for all test functions $v$ yields the
minimality condition
\begin{equation}
\label{RLEL}
\left(H^* \circledast \left(1-\frac{f}{H\circledast u}\right)\right)u = 0\;.
\end{equation}
Because of the energy conservation property $H^*\circledast 1=1$ one 
sees that \eqref{RL} is a fixed point iteration for \eqref{RLEL}.

\subsection{Regularisation}

In the presence of noise the functional \eqref{RLfunc} is not
minimised by a smooth function $u$; in fact, the fixed-point iteration 
shows the above-mentioned semi-convergence behaviour and
diverges for $k\to\infty$. From the variational viewpoint,
the functional \eqref{RLfunc} needs to be regularised. 
In standard Richardson-Lucy deconvolution, this regularisation is provided 
implicitly by stopping the iteration after a finite number of steps. 
The earlier the iteration is stopped, the higher is the degree of 
regularisation.
Although this sort of regularisation is not represented in the
functional \eqref{RLfunc},
the variational picture is advantageous because \eqref{RLfunc} can be modified
in the same flexible way as standard variational approaches.
The structure of the iterative minimisation procedure is preserved
throughout these modifications, which leads to good computational efficiency.

Let us first note that
by limiting the growth of high-frequency signal components, regularisation
has a smoothing effect that in deconvolution problems acts contrary to
the intended image sharpening. It is desirable to steer this
effect in such a way that it interferes as little as
possible with the enhancement of salient image structures, such as edges.

Implicit regularisation by stopping, however, allows little control
over the way it affects image structures.
For this reason, it makes sense to introduce a variational regularisation
term into the objective functional. This yields the functional
\begin{equation}
\label{NRRLfunc}
E_{f,H}[u] = \int\limits_{\Omega} \left(
r_f(H\circledast u)
+ \alpha\, \varPsi(\lvert\nabla u\rvert^2)
\right)
\dd\bx
\end{equation}
in which the Richardson-Lucy data term is complemented by a regulariser
whose influence is weighted by the regularisation weight $\alpha>0$.
Concerning the penalisation function $\varPsi(\argdot)$ in the regulariser,
our discussion from Section~\ref{sec-exdcvm} applies analogously.
With the total variation regulariser given by $\varPsi(z^2)=\lvert z\rvert$,
the energy functional \eqref{NRRLfunc} corresponds to
the method proposed (in space-invariant formulation) by Dey et
al.\ \cite{Dey-isbi04,Dey-MRT06}; compare also the more recent work 
\cite{Sawatzky-aip10} for a similar approach.

The Euler-Lagrange equation for \eqref{NRRLfunc} under multiplicative
perturbation is given by
\begin{equation}
0 = \biggl(H^* \circledast \left(1-\frac{f}{H\circledast u}\right) 
- \alpha\, \Div \bigl(\varPsi'(\lvert\nabla u\rvert^2)\,\nabla u\bigr)
\biggr) \cdot u \;,
\label{NRRLEL}
\end{equation}
which combines \eqref{RLEL} with the 
same 
multiplicative gradient for the regulariser
as in \eqref{VGDphi},
compare \cite{Welk-ibpria07}.

In converting this into a fixed point iteration, we evaluate the divergence 
expression with $u=u^k$, yielding
$D^k:=\alpha \Div \left(\varPsi'(\lvert\nabla u^k\rvert^2)\,\nabla u^k
\right)$. Dependent on whether the factor $u$ with which the divergence 
term in \eqref{NRRLEL} is multiplied is chosen as $u^k$ or $u^{k+1}$,
the right-hand side of \eqref{RL} either receives the additional summand 
$D^k$, or is divided by $(1-D^k)$. However, $D^k$ can have either 
sign, and a negative value in the numerator or denominator will lead to
a violation of the positivity requirement.
For this reason, we choose the outer factor for $D^k$ as $u^k$ if $D^k>0$,
or $u^{k+1}$ if $D^k<0$. Using the abbreviations 
$[z]_\pm:=\frac12(z\pm\lvert z\rvert)$
we can therefore write our final fixed point iteration as
\begin{equation}
u^{k+1} = 
\frac
{H^*\circledast \left(\frac{f}{H\circledast u^k}\right) + 
\alpha \left[\Div\bigl(
\varPsi'(\lvert\nabla u^k\rvert^2)\,\nabla u^k\bigr)\right]_+}
{1 - 
\alpha \left[\Div\bigl(
\varPsi'(\lvert\nabla u^k\rvert^2)\,\nabla u^k\bigr)\right]_-}
u^k\;.
\label{NRRL}
\end{equation}
We will refer to this method as \emph{regularised RL}.

\subsection{Robust Data Terms}

Up to scaling and shifting, the asymmetric penaliser function $r_f(w)$ 
equals the logarithmic density of a Gamma distribution.
Minimisation of the integral \eqref{RLfunc} thus corresponds to
a Bayesian estimation of the sharp image assuming a Poisson distribution
for the intensities, with the Gamma distribution as conjugate prior.

In 
variational deconvolution, it has turned out useful
to replace quad\-ra\-tic data terms that mirror a Gaussian noise model 
by robust data terms associated with ``heavy-tailed'' noise distributions. 
Not only can the resulting model handle extreme noise but it can also 
cope with imprecisions in the blur model.
Following this idea, we replace the data term of \eqref{NRRLfunc}
by one that is adapted to a broader distribution on $\real^+$.
To retain the structure of the fixed point iterations \eqref{RL}
and \eqref{NRRL}, we 
keep $r_f(H\circledast u)$ in the data term, but apply a 
penaliser function $\varPhi$ that grows less than linear.

Our modified functional therefore reads
\begin{equation}
\label{RRRLfunc}
E_{f,H}[u] = \int\limits_\Omega \left(
\varPhi\bigl(r_f(H\circledast u)\bigr)
+ \alpha\, \varPsi(\lvert\nabla u\rvert^2)
\right)
\dd\bx \;.\kern-.7em
\end{equation}

By an analogous derivation as before, one obtains for \eqref{RRRLfunc}
the minimality condition
\begin{align}
\label{RRRLEL}
0 &= \biggl(H^* \circledast \left( \varPhi'\bigl(r_f(H\circledast u)\bigr)
\left(1-\frac{f}{H \circledast u}\right) \right) 
- \alpha\,\Div\bigl(\varPsi'(\lvert\nabla u\rvert^2)\,\nabla u\bigr)
\biggr) \cdot u 
\end{align}
that leads to 
the new fixed point iteration
\begin{equation}
u^{k+1} = 
\frac
{
H^*\circledast\left(\varPhi'\bigl(r_f(H\circledast u)\bigr)\,
\frac{f}{H\circledast u^k}
\right)+ 
\alpha \left[\Div\bigl(
\varPsi'\bigl(\lvert\nabla u^k\rvert^2\bigr)\,\nabla u^k\bigr)\right]_+
}
{
H^*\circledast\varPhi'(r_f(H\circledast u)) 
- 
\alpha \left[\Div\bigl(
\varPsi'(\lvert\nabla u^k\rvert^2)\,\nabla u^k\bigr)\right]_-
}
\,\cdot\,
u^k
\label{RRRL}
\end{equation}
which we call \emph{robust and regularised RL deconvolution} (RRRL).
Comparing to \eqref{NRRL}, computational cost grows by one more
convolution and the evaluation of $\varPhi'$. 

Clearly, \eqref{RRRL} contains regularised RL \eqref{NRRL} as 
special case ($\varPhi'\equiv 1$). Similarly, $\alpha=0$ yields a 
non-regularised method which we will call \emph{robust RL deconvolution}.

\subsection{Multi-Channel Images}

Assume now that the blurred image is a multi-channel image 
$\bmf=(f_i)_{i\in\varGamma}$ with a channel index set $\varGamma$,
e.g.\ an RGB colour image, whose channels are uniformly blurred, i.e.\
the PSF $H$ is equal for all channels.
Replacing 
the expressions $r_f(H\circledast u)$
and $\lvert\nabla u\rvert^2$ in the arguments of $\varPhi$ and $\varPsi$ 
with their sums over image channels,
$R=\sum\limits_{i\in \varGamma}r_{f_i}(H\circledast u_i)$ and
$G=\sum\limits_{i\in \varGamma}\lvert\nabla u_i\rvert^2$, we obtain
as multi-channel analog of \eqref{RRRL} the functional
\begin{equation}
\label{RRRLMCfunc}
E_{\bmf,H}[\bu] = \int\limits_\Omega \bigl(
\varPhi(R) 
+ \alpha\, \varPsi(G)
\bigr)
\dd\bx
\;.
\end{equation}

This yields as the iteration rule for multi-channel RRRL
\begin{equation}
u_j^{k+1} =
\frac
{
H^*\circledast\left(\varPhi'(R)\, \frac{f_j}{H\circledast u_j^k}\right) 
+
\alpha \left[\Div\left( \varPsi'(G)\, \nabla u_j^k\right)\right]_+
}
{
H^*\circledast\varPhi'(R) - \alpha \left[\Div\left( \varPsi'(G)\,
\nabla u_j^k\right)\right]_-
}
\,\cdot\,
u_j^k\;.
\label{RRRLMC}
\end{equation}
The same procedure works for the non-robust and/or non-regularised RL variants.
Note that in the case of standard RL this boils down to channel-wise 
application.

\subsection{Stability}\label{ssec-stab}

The regularised RL model by Dey et al.\ \cite{Dey-MRT06} places the
regularisation only in the denominator of the fixed point iteration
rule. This imposes a tight bound on $\alpha$: 
as soon as
$\alpha\,\Div\left( \varPsi'(G)\,\nabla u_j^k\right)$ exceeds $1$,
positivity is violated, and the iteration becomes unstable, 
In our iteration rule, sign-dependent distribution of
divergence expressions to the enumerator and denominator prevents positivity
violations, thus enabling larger values of $\alpha$.

Nevertheless, for substantially larger values of $\alpha$ instabilities
are observed which can be attributed to amplifying
perturbations of high spatial frequency in almost homogeneous image regions.
In \cite{Persch-MST13x} therefore a modification of the fixed point
iteration to optimise \eqref{RRRLfunc} has been proposed that guarantees
stability for arbitrary $\alpha$.
A detailed analysis of stability bounds on $\alpha$ in the
iteration \eqref{RRRL} will be given in a forthcoming publication.

\section{Experiments}\label{sec-exp}

We turn now to evaluate the performance of our newly developped deconvolution 
methods, and comparing them to existing approaches. Methods chosen for
comparison include classical Richardson-Lucy deconvolution, variational
gradient descent methods, and the methods from 
\cite{Krishnan-nips09,Wang-SIIMS08} which are advocated for performant 
deblurring in several recent papers, see e.g.\ \cite{Hirsch-iccv11}.

\subsection{Deconvolution of Synthetic Data} 

We start our tests on
grey-value images that are synthetically blurred by convolution.
Since in this case the correct sharp image is known, we can rate restoration 
quality by the signal-to-noise ratio 
\begin{equation}
\label{SNR}
\mathrm{SNR}(u,g) = 10~\mathrm{log}_{10}\frac{\mathrm{var}(g)}
{\mathrm{var}(g-u)}\,\mathrm{dB}
\;.
\end{equation}
Here, $g$ and $u$ are the original sharp image and restored image, 
respectively.
By $\mathrm{var}(v)$ we denote the variance of the image $v$.
One should be aware that SNR measurements do often not capture
visual quality of deblurred images very well. The parameters in our
experiments are optimised primarily for visual quality, not for SNR.

We remark also that in synthetically blurring images, we use convolution
via the Fourier domain, which involves treating the image domain as periodic.
In contrast, we use convolution in the spatial domain in the variational 
deconvolution procedures as well as in the Richardson-Lucy method and its 
modifications. 
This discrepancy in the 
convolution procedure and boundary treatment is by purpose: it helps
to prevent ``inverse crimes'' \cite{Colton-Book92} that could unduly 
embellish results. Moreover, our implementations can thereby easily be
adapted to spatially variant blurs.

The methods from \cite{Krishnan-nips09,Wang-SIIMS08} 
require computations in the Fourier transforms by design. 

\begin{figure}[t!]
\centering
\begin{tabular}{@{}c@{~}c@{~}c@{~}c@{~}c@{}}
\includegraphics[width=0.19\textwidth]{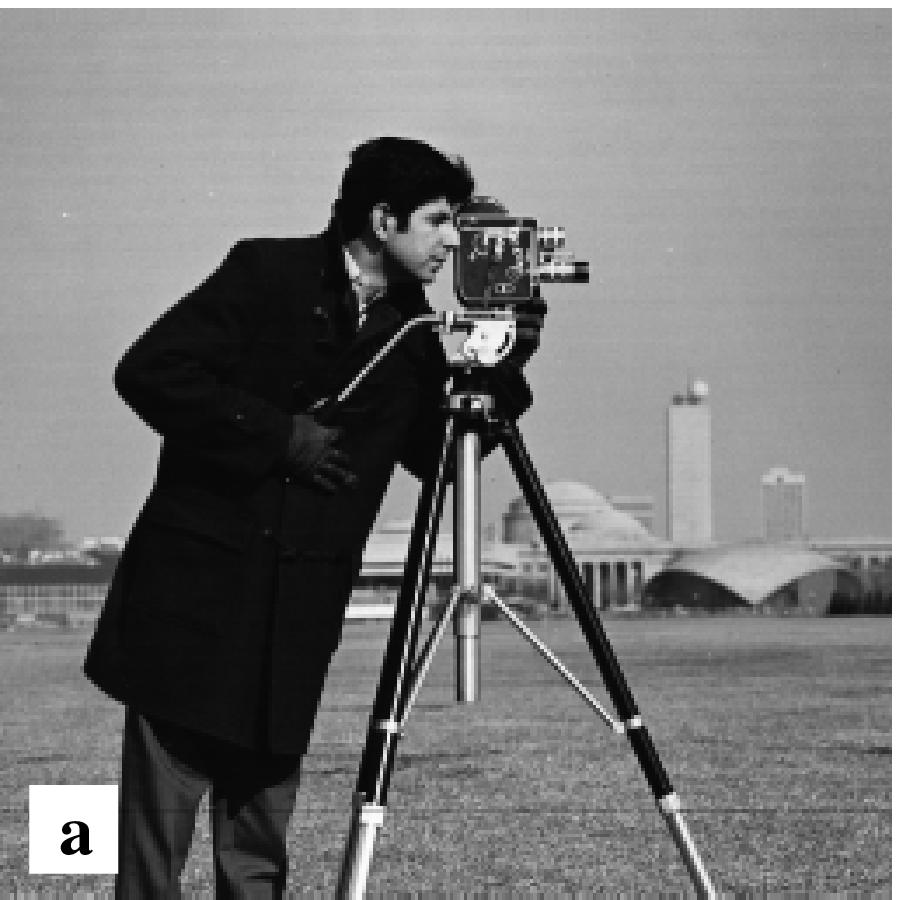}&
\includegraphics[width=0.19\textwidth]{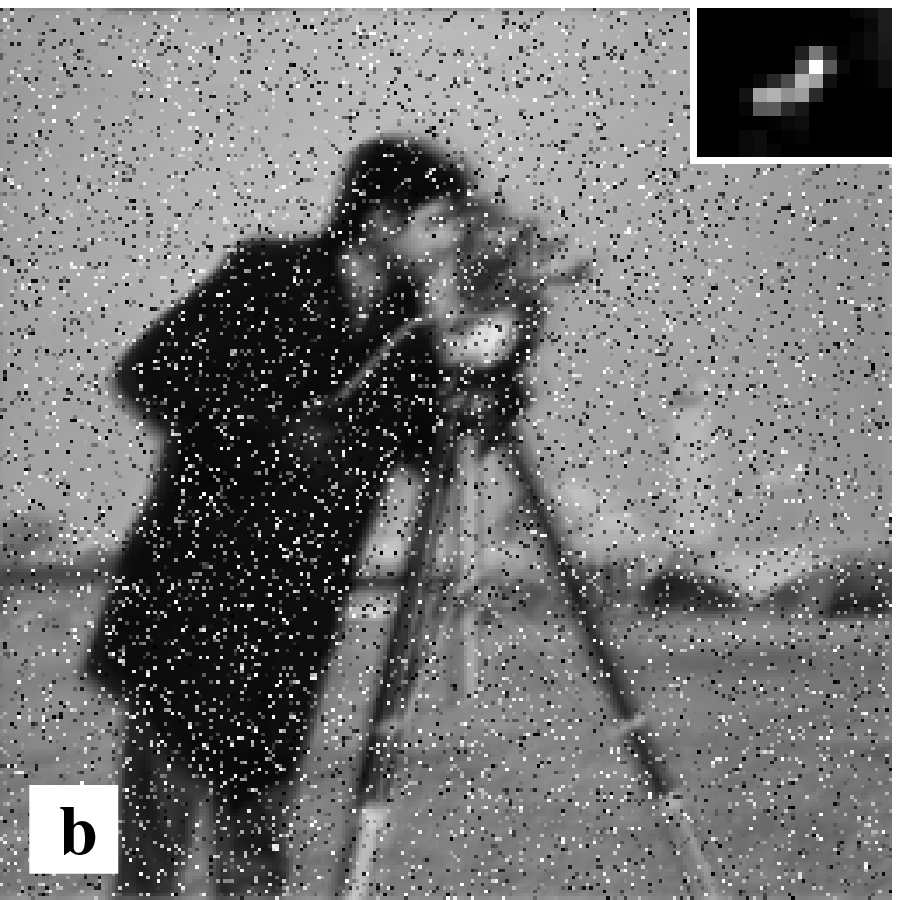}&
\includegraphics[width=0.19\textwidth]{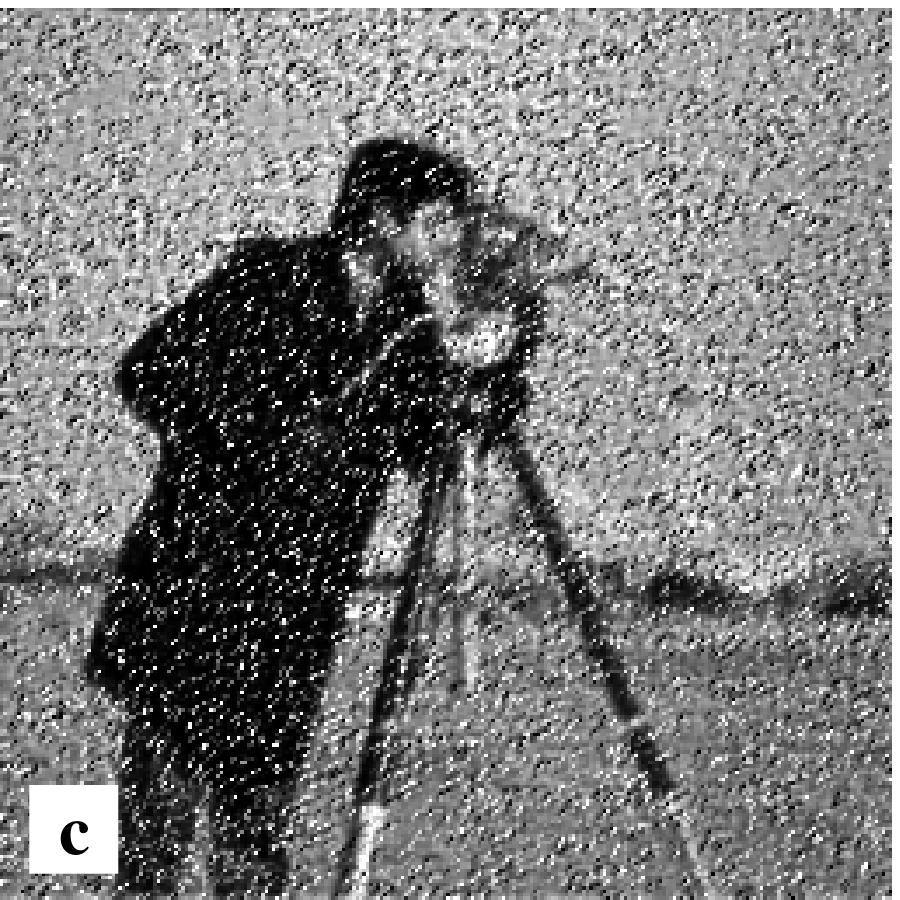}&
\includegraphics[width=0.19\textwidth]{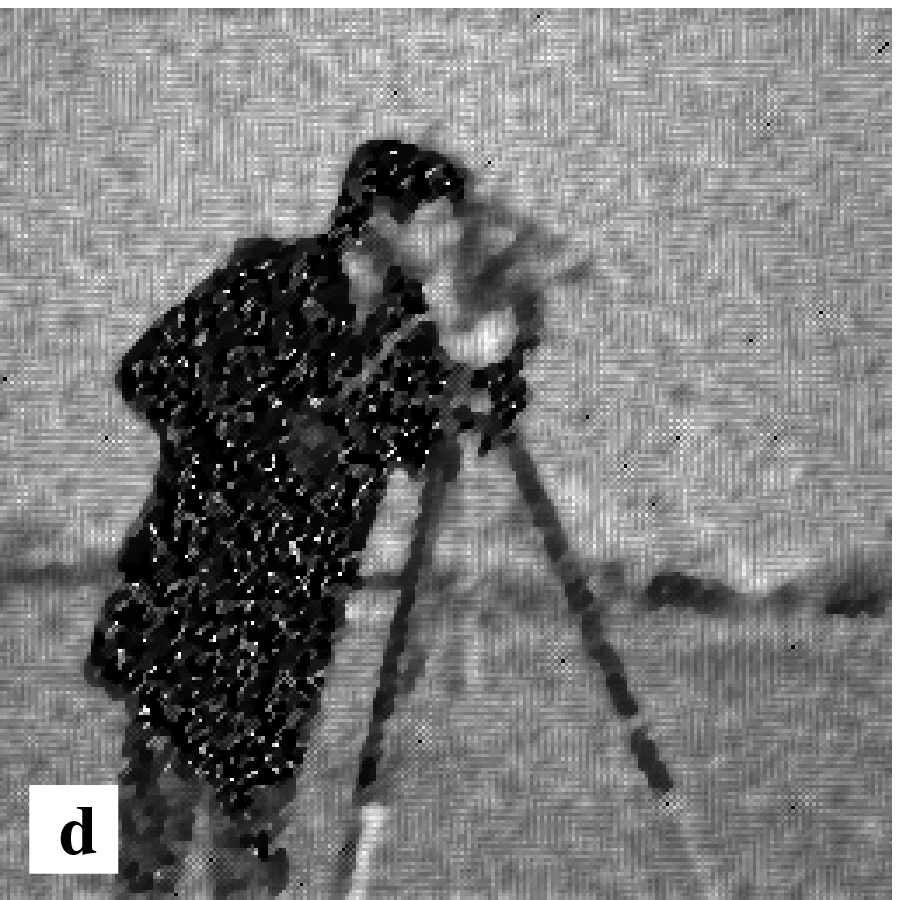}&
\includegraphics[width=0.19\textwidth]{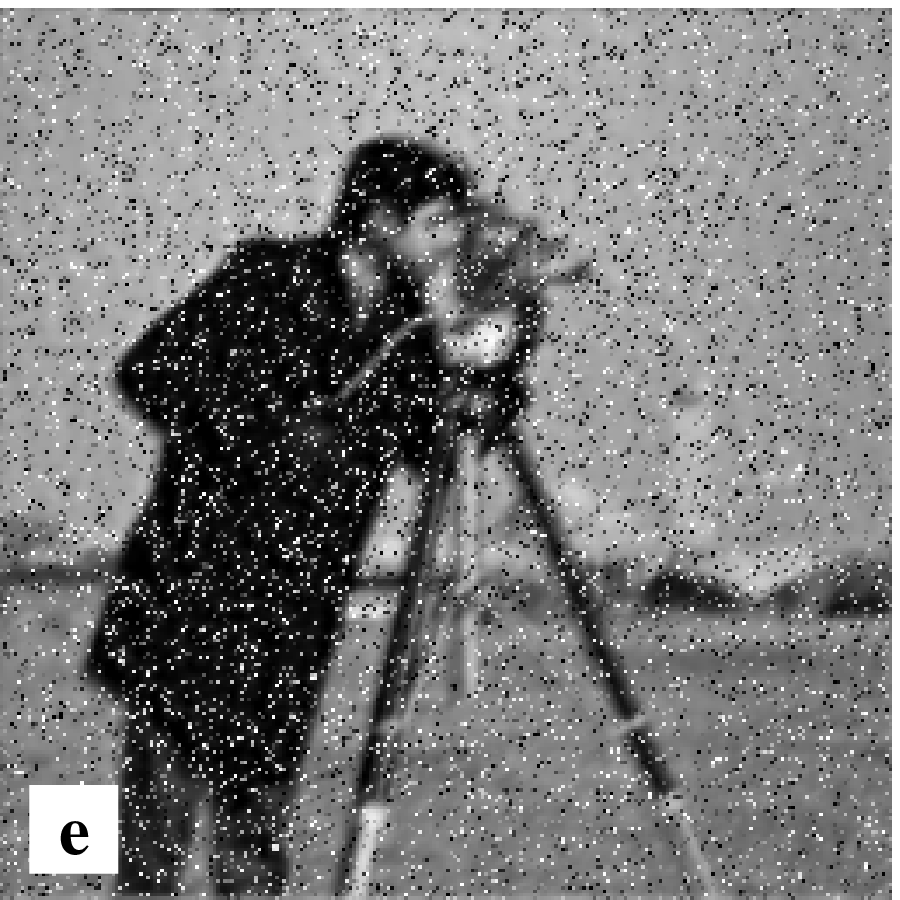}\\
\includegraphics[width=0.19\textwidth]{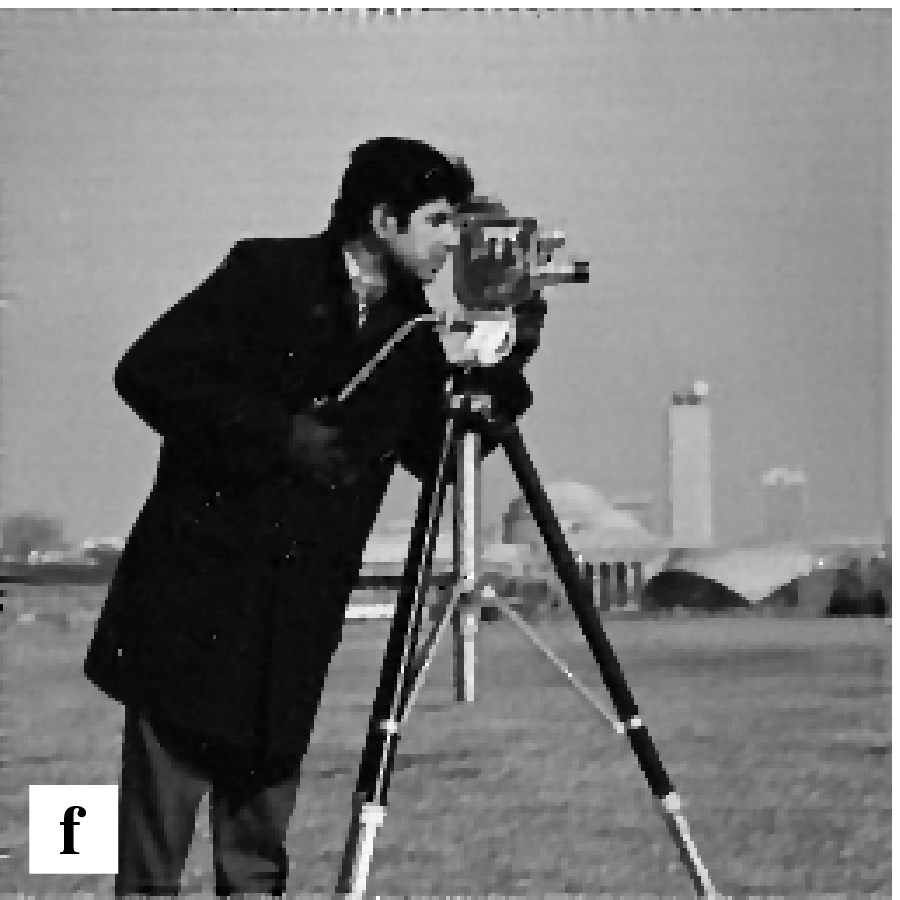}&
\includegraphics[width=0.19\textwidth]{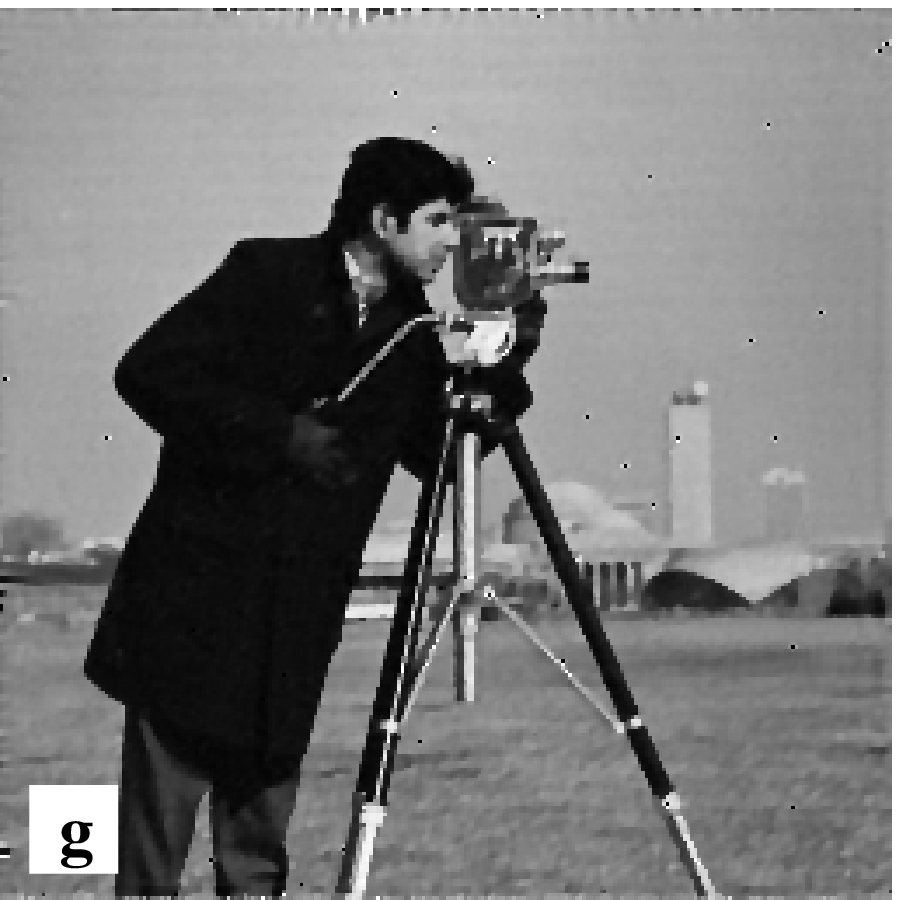}&
\includegraphics[width=0.19\textwidth]{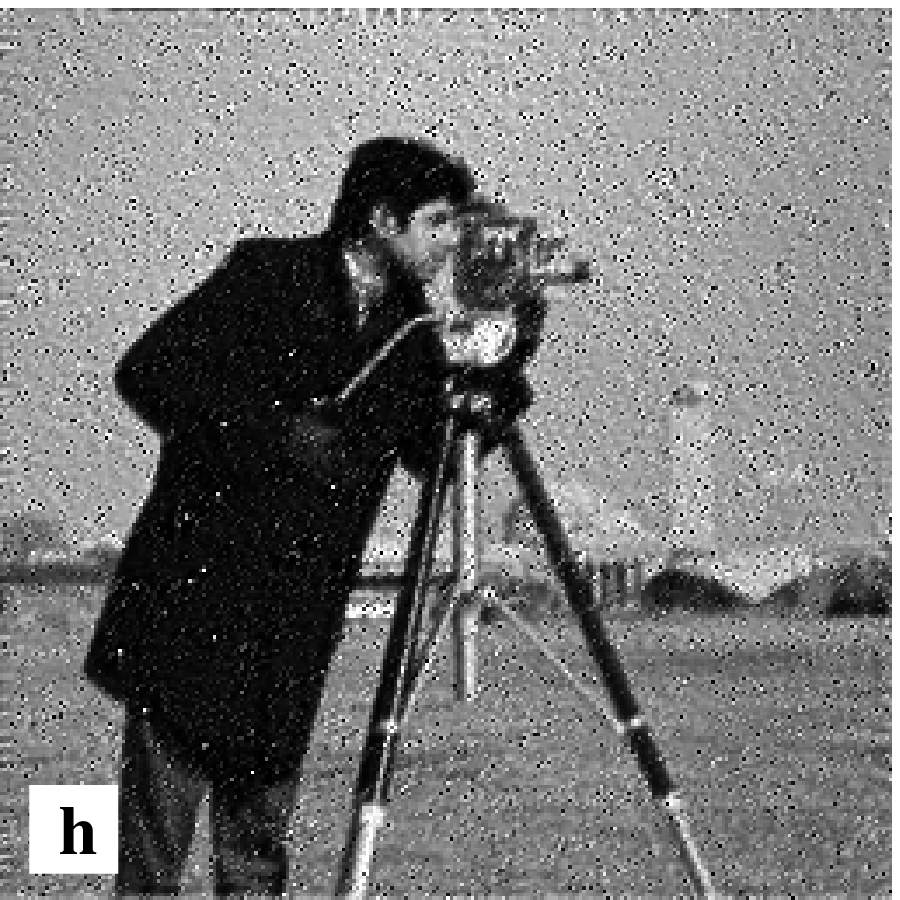}&
\includegraphics[width=0.19\textwidth]{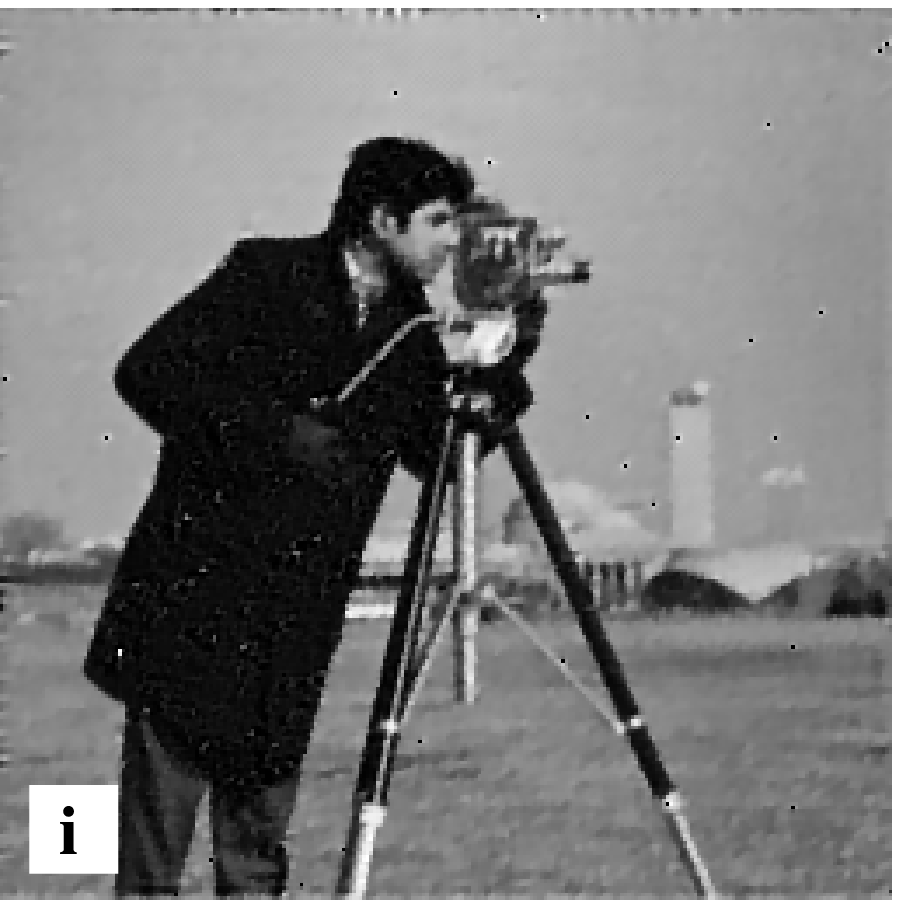}&
\includegraphics[width=0.19\textwidth]{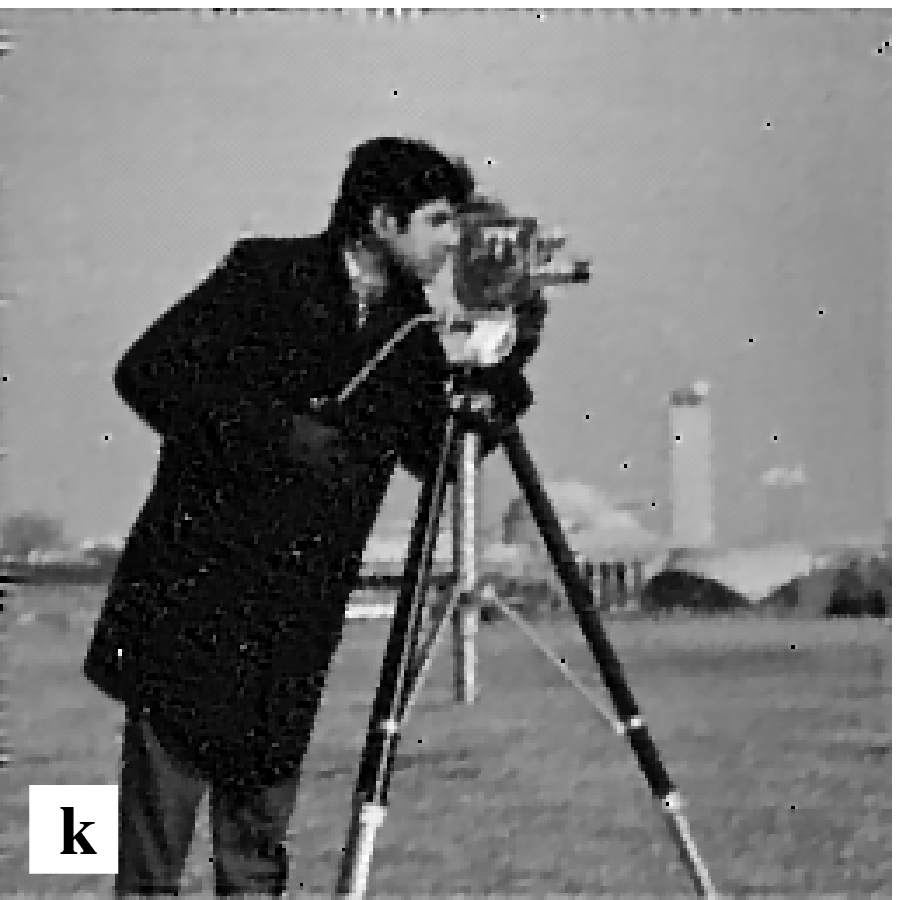}
\end{tabular}
\caption{\label{f-sy1-sivp}%
\textbf{(a)}~\emph{Cameraman} image, $256\times256$ pixels.
\textbf{(b)}~Moderately blurred, and $15\,\%$ of all pixels replaced by
uniform noise; SNR: $3.66\,\mathrm{dB}$. 
Insert shows PSF (four times enlarged).
\textbf{(c)}~Image from (b) 
deblurred by standard RL \eqref{RL}, $10$ iterations;
SNR: $2.28\,\mathrm{dB}$.
\textbf{(d)}~Regularised RL \eqref{NRRL}, $\alpha=0.1$, 
$100$ iterations;
SNR: $6.40\,\mathrm{dB}$.
\textbf{(e)}~Deblurred by the method from \cite{Krishnan-nips09},
$\lambda=1$, $\beta$ from $1$ to $256$ with step factor $2\sqrt2$,
$1$ iteration per level,
SNR: $3.62\,\mathrm{dB}$.
\textbf{(f)}~Robust variational deconvolution without constraints
\cite{Welk-dagm05} using $L^1$ data term and Perona-Malik regulariser
($\lambda=15$), regularisation weight $\alpha=0.06$;
SNR: $16.03\,\mathrm{dB}$.
\textbf{(g)}~Same but with positivity
constraint, see \cite{Welk-ibpria07} and Section~\ref{sec-rrrl},
regularisation weight $\alpha=0.06$; 
SNR: $15.42\,\mathrm{dB}$.
\textbf{(h)}~Robust RL, $50$ iterations;
SNR: $6.42\,\mathrm{dB}$. 
\textbf{(i)}~RRRL \eqref{RRRL}
with regularisation weight $\alpha=0.005$, $200$ iterations; 
SNR: $14.41\,\mathrm{dB}$.
\textbf{(k)}~Same but $2000$ iterations; SNR: $14.21\,\mathrm{dB}$.
}
\end{figure}

\begin{figure}[t!]
\centering
\begin{tabular}{@{}c@{~}c@{~}c@{~}c@{}}
\includegraphics[width=0.19\textwidth]{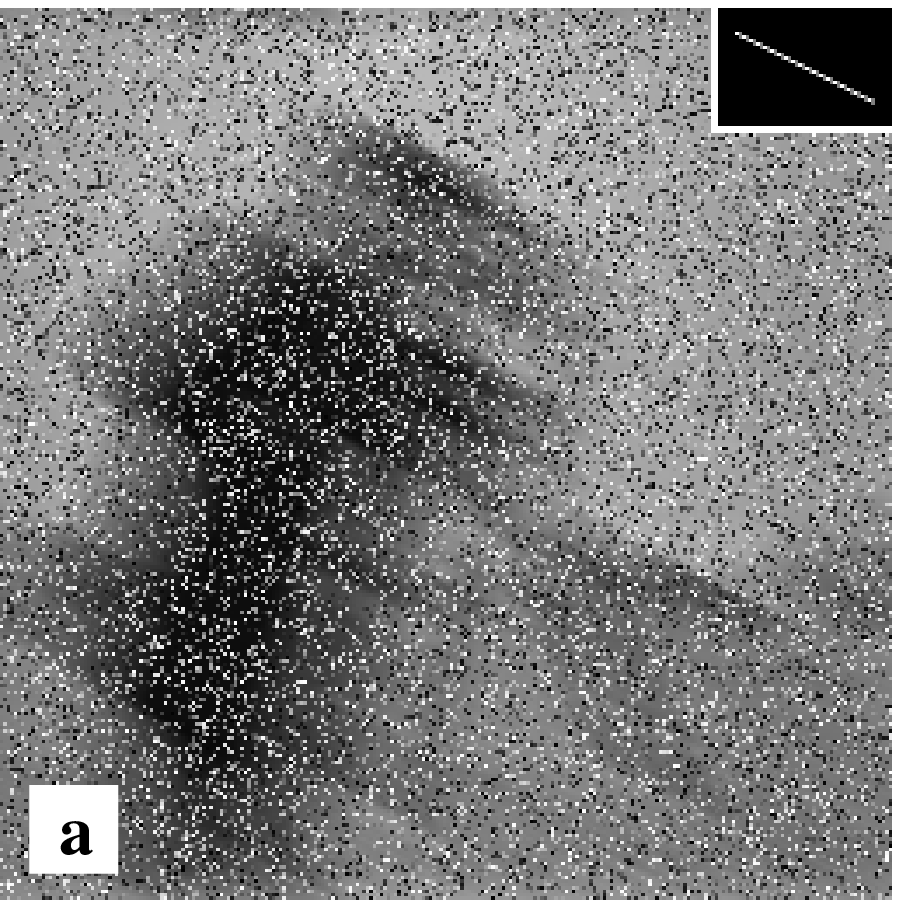}&%
\includegraphics[width=0.19\textwidth]{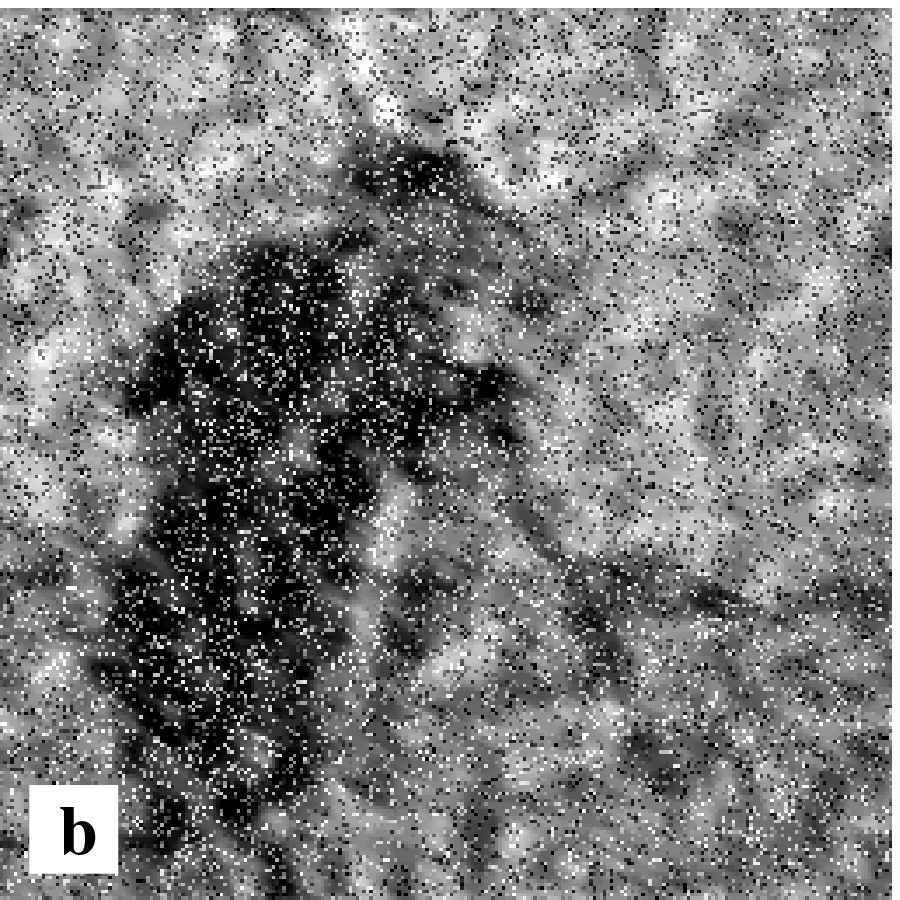}&%
\includegraphics[width=0.19\textwidth]{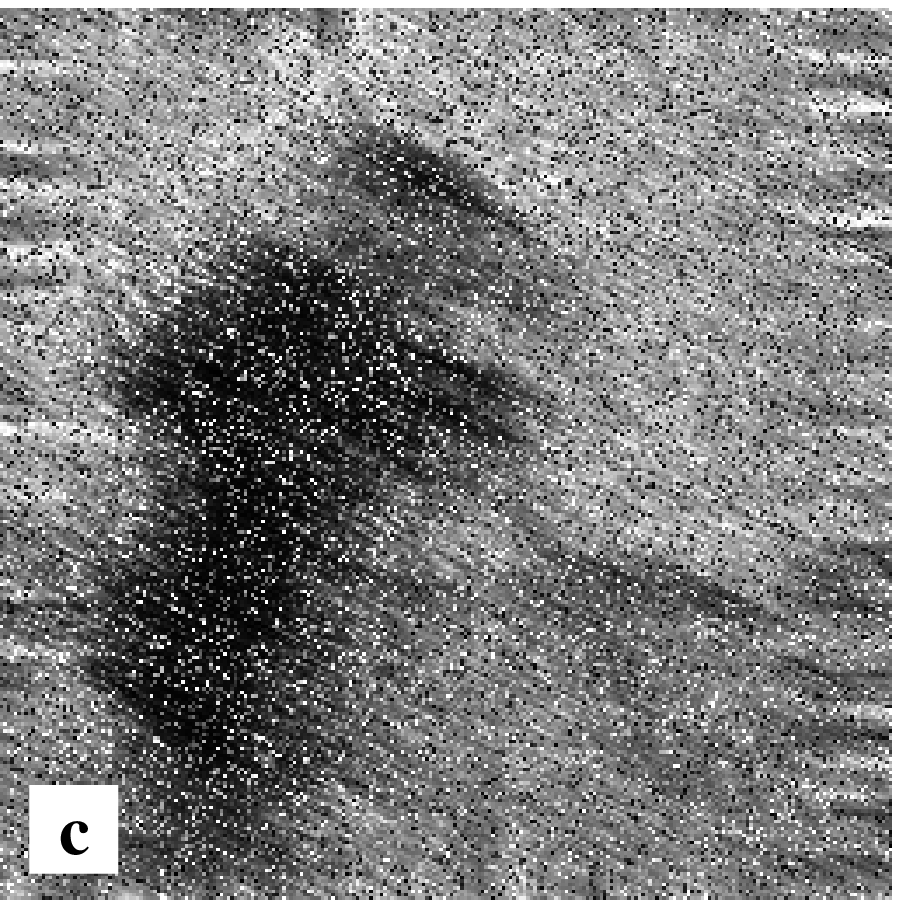}&%
\includegraphics[width=0.19\textwidth]{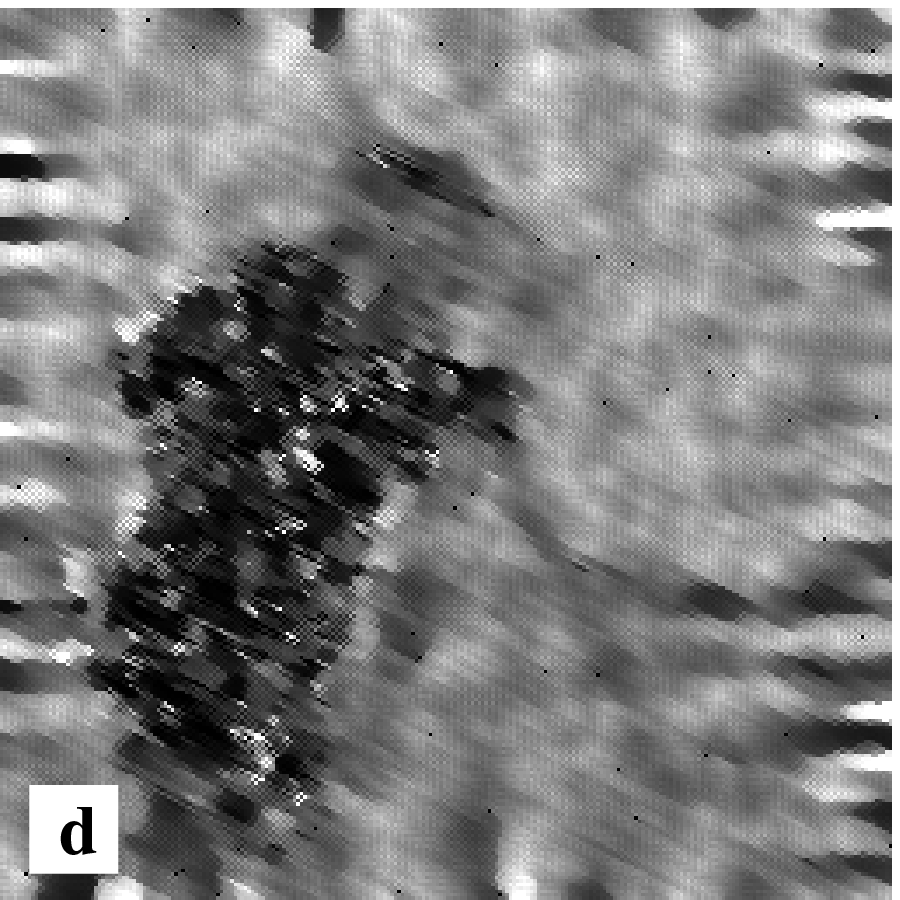}
\end{tabular}
\caption{\label{f-sy2a}%
\textbf{(a)}~\emph{Cameraman} image from Figure~\ref{f-sy1-sivp}(a)
severely blurred, and $30\,\%$ of all pixels 
replaced by uniform noise, signal-to-noise ratio: $0.26\,\mathrm{dB}$. 
Insert shows PSF (true size).
\textbf{(b)}~Deblurred by the method from \cite{Krishnan-nips09},
$\lambda=0.5$, $\beta$ from $1$ to $256$ with step factor $\sqrt2$,
$5$ iterations per level,
SNR: $0.52\,\mathrm{dB}$.
\textbf{(c)}~Deblurred by RL \eqref{RL}, $10$ iterations,
SNR: $0.09\,\mathrm{dB}$.
\textbf{(d)}~Regularised RL \eqref{NRRL}, $100$ iterations,
with regularisation weight $\alpha=0.05$;
SNR: $2.32\,\mathrm{dB}$.
}
\end{figure}

\begin{figure}[t!]
\centering
\begin{tabular}{@{}c@{~}c@{~}c@{~}c@{}}
\includegraphics[width=0.19\textwidth]{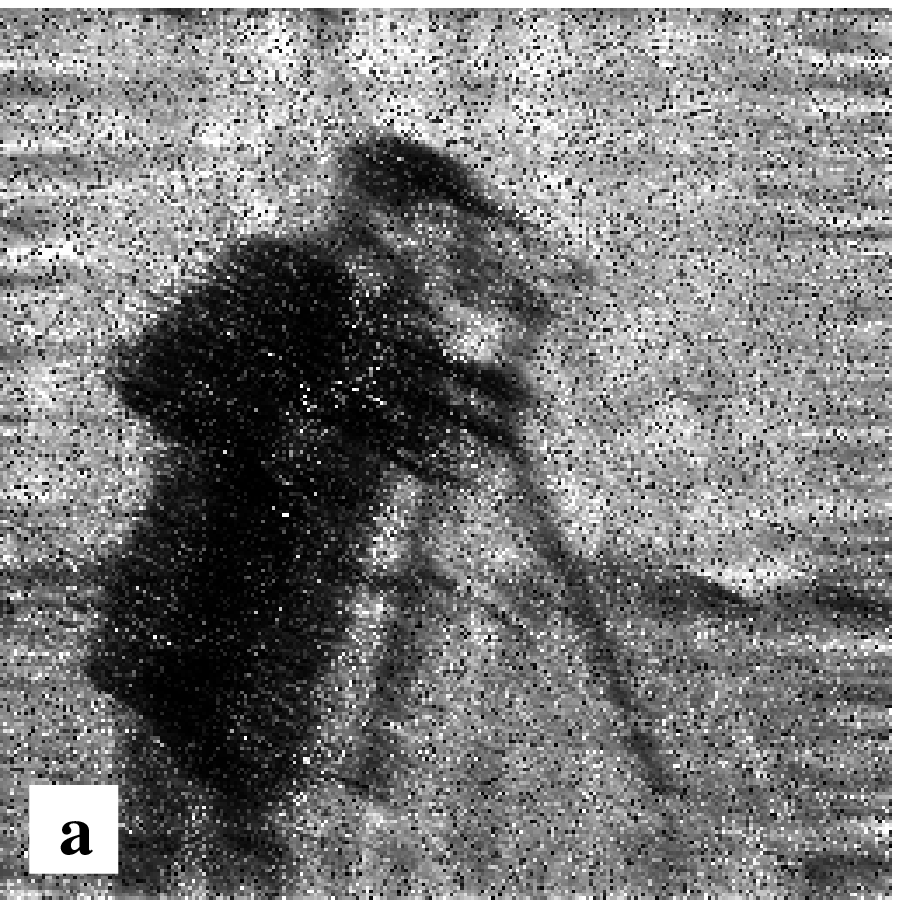}&%
\includegraphics[width=0.19\textwidth]{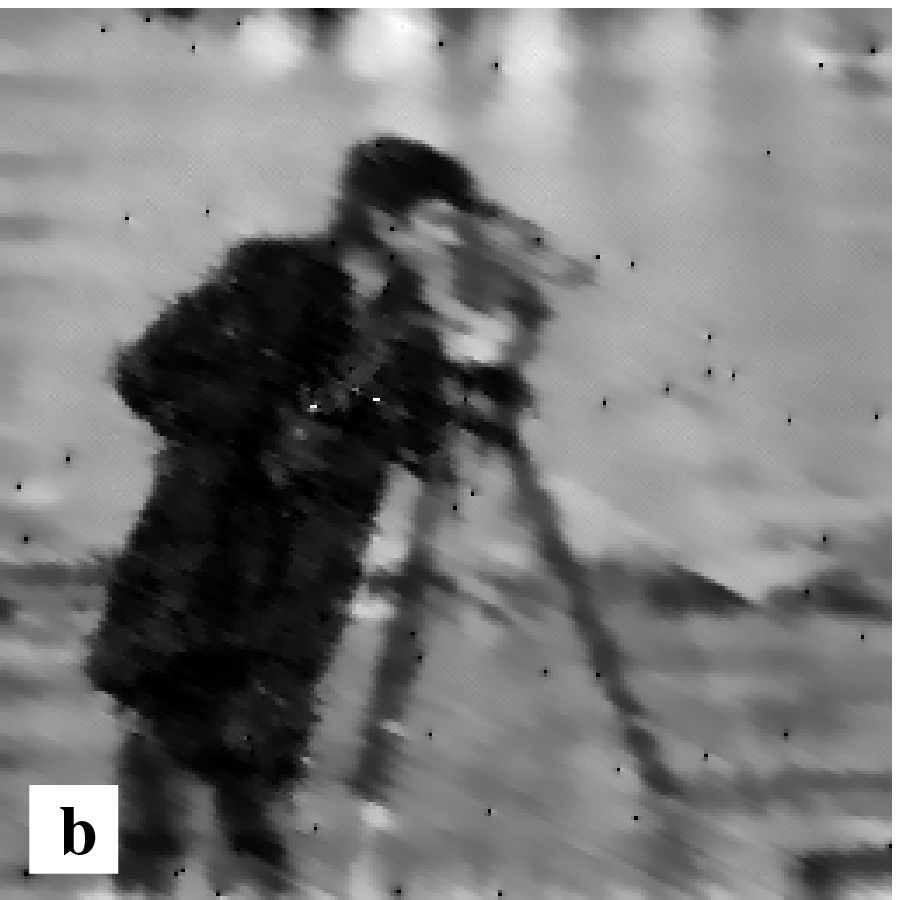}&%
\includegraphics[width=0.19\textwidth]{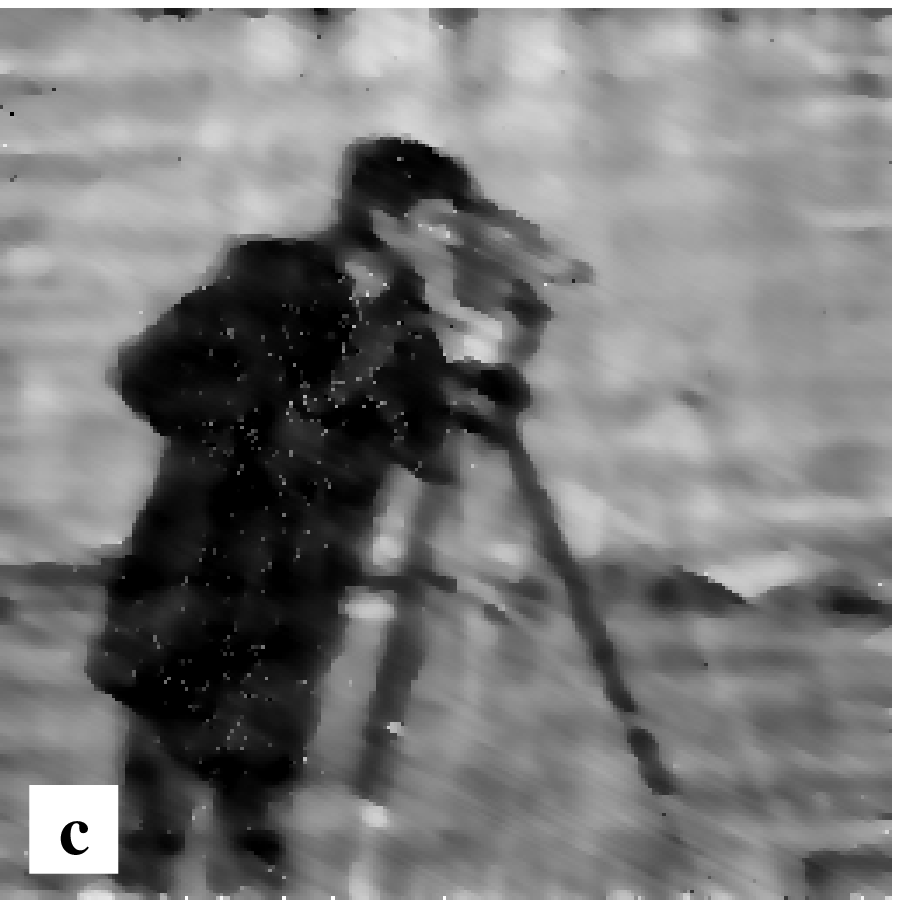}&%
\includegraphics[width=0.19\textwidth]{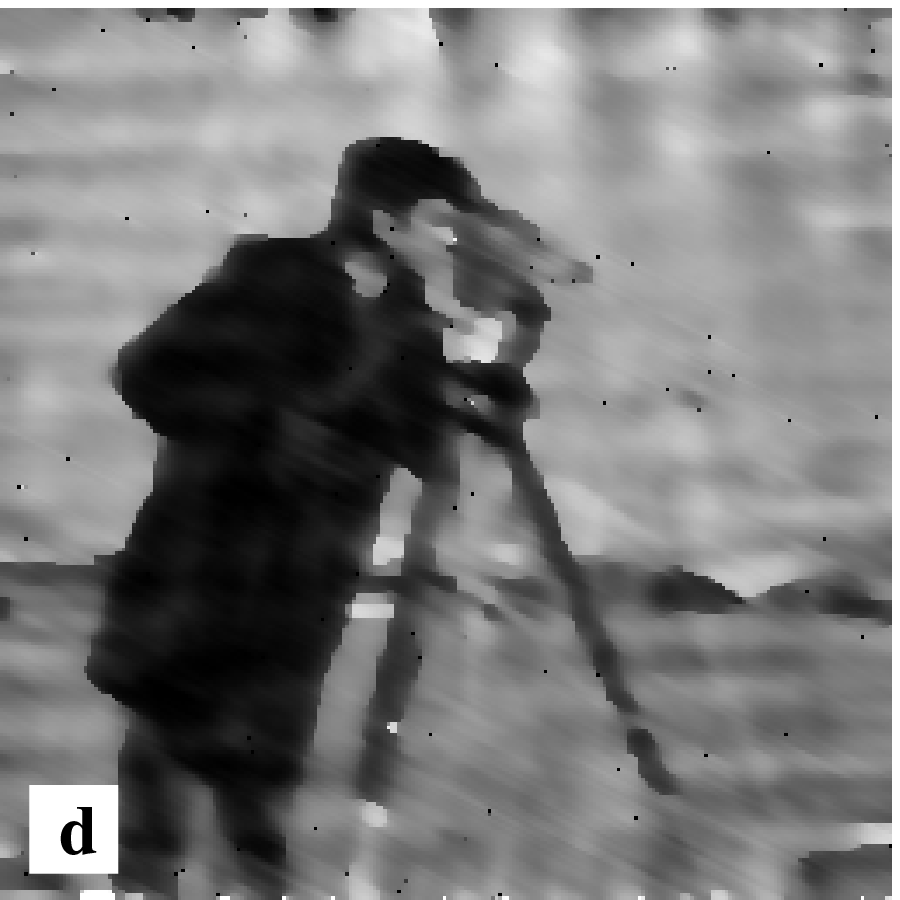}
\end{tabular}
\caption{\label{f-sy2b}%
\textbf{(a)}~Image from Figure~\ref{f-sy2a}(b) deblurred by
robust RL, $100$ iterations, 
SNR: $1.84\,\mathrm{dB}$.
\textbf{(b)}~RRRL \eqref{RRRL} with 
regularisation weight $\alpha=0.003$, $400$ iterations; 
SNR: $7.42\,\mathrm{dB}$.
\textbf{(c)}~Robust variational deconvolution without constraint,
$\alpha=0.06$; SNR: $7.52\,\mathrm{dB}$. 
\textbf{(d)} Robust variational deconvolution with positivity 
constraint, $\alpha=0.09$; SNR: $7.60\,\mathrm{dB}$.
}
\end{figure}

\paragraph{First series of experiments.}
In this series of experiments 
(Figure~\ref{f-sy1-sivp}),
we blur the \emph{cameraman} image, 
Figure~\ref{f-sy1-sivp}(a),
by an irregularly shaped point-spread function 
of moderate size and apply impulsive (uniform) noise (b). 
The so degraded 
image can be deblurred to some extent by standard RL, 
(c),
but noise is severely amplified, and dominates the result when 
the iteration count is increased for further 
sharpening.
Regularised RL 
visibly reduces the 
noise effect, see 
frame~(d).

Turning to classical variational approaches, we consider in 
frame~(e)
the iterative method from \cite{Krishnan-nips09}. 
Minimising an energy functional with quadratic data term and an edge-enhancing
regulariser, it still suffers from severe noise effects in our example.
The closely related method from \cite{Wang-SIIMS08} as well as other
TV deconvolution approaches lead to similar results.

Robust variational deconvolution achieves a significantly better
restoration, see 
frame~(f)
where the method from \cite{Welk-dagm05} without positivity constraint is 
used, and 
frame~(g)
with positivity constraint as in \cite{Welk-ibpria07} and 
Section~\ref{sec-rrrl}.
Visually, both variational approaches lead to a comparable
restoration quality. 
In
this example the
positivity constraint bears no improvement,
due to the fact that
artifact suppression by the positivity constraint takes effect mainly in dark
image regions which play only a minor role in the cameraman image. 

Using robust RL without regularisation as in 
frame~(h)
reduces the noise to an extent comparable to the regularised method in 
(d).
With more iterations, 
however, the noise still becomes dominant, thus limiting the possible 
deconvolution quality. 
Robust and regularised RL allows fairly good deblurring while only small 
rudiments of noise remain, see 
Figure~\ref{f-sy1-sivp}(i).
Even if the iteration count is drastically increased, such that the
implicit regularisation by stopping takes only little effect, the explicit
regularisation remains effective 
(k)
and keeps the noise level low.
Indeed, the reconstruction quality comes close, both visually and in terms
of SNR, to that achieved by state-of-the-art robust variational 
deconvolution as shown before in 
(f) and (g).

\paragraph{Second series of experiments.}
Encouraged by the previous findings, we increase blur and noise in our 
second series of experiments (Figures~\ref{f-sy2a}--\ref{f-sy2b}). Under the 
influence of a drastically larger simulated motion blur and doubled noise 
intensity, see Figure~\ref{f-sy2a}(a),
classic RL gives no longer usable results~(c). The same is true for
the method from \cite{Krishnan-nips09}, see Subfigure~(b).
Regularised RL and robust RL can again cope better with the noise but 
their outcomes are far from satisfactory, see Figure~\ref{f-sy2a}(d) and
\ref{f-sy2b}(a). A significant improvement is obtained with
robust and regularised RL, Figure~\ref{f-sy2b}(b), as well as with 
robust variational deconvolution without constraints (c) or with
positivity constraint (d).
Although the restoration quality is still imperfect, the three last mentioned
methods are eye to eye in terms of visual quality and SNR.

\begin{figure}[!]
\centering
\begin{tabular}{@{}c@{~}c@{}}
\includegraphics[width=0.24\textwidth]{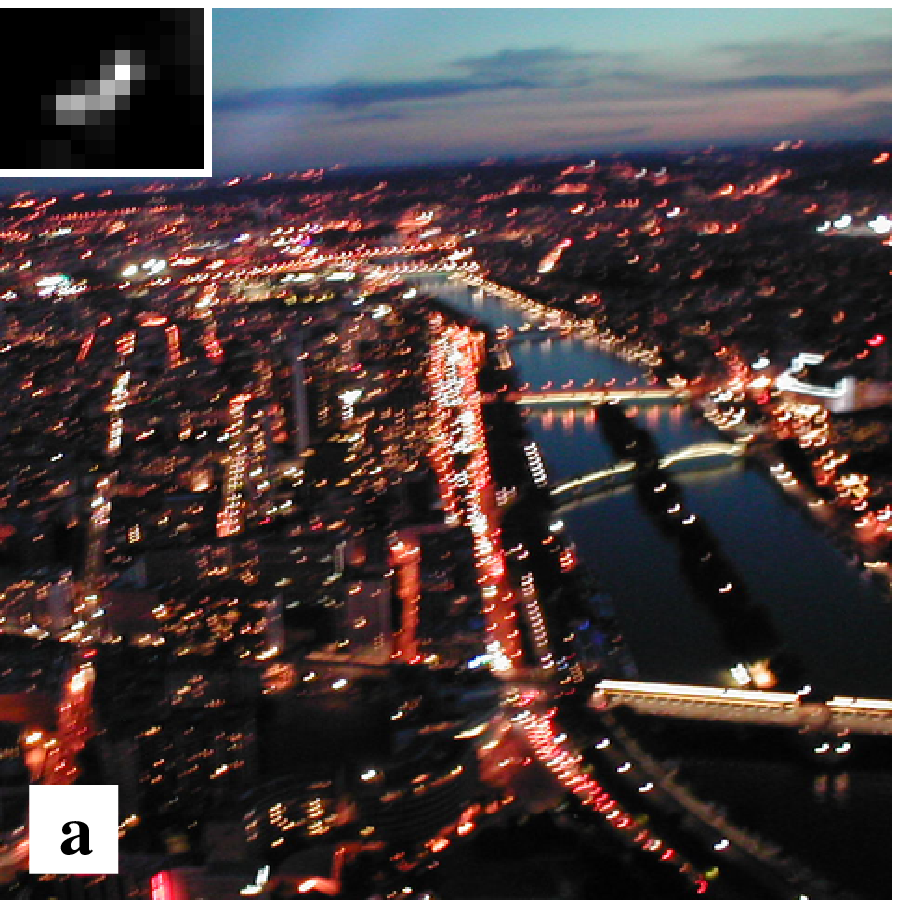}&
\includegraphics[width=0.24\textwidth]{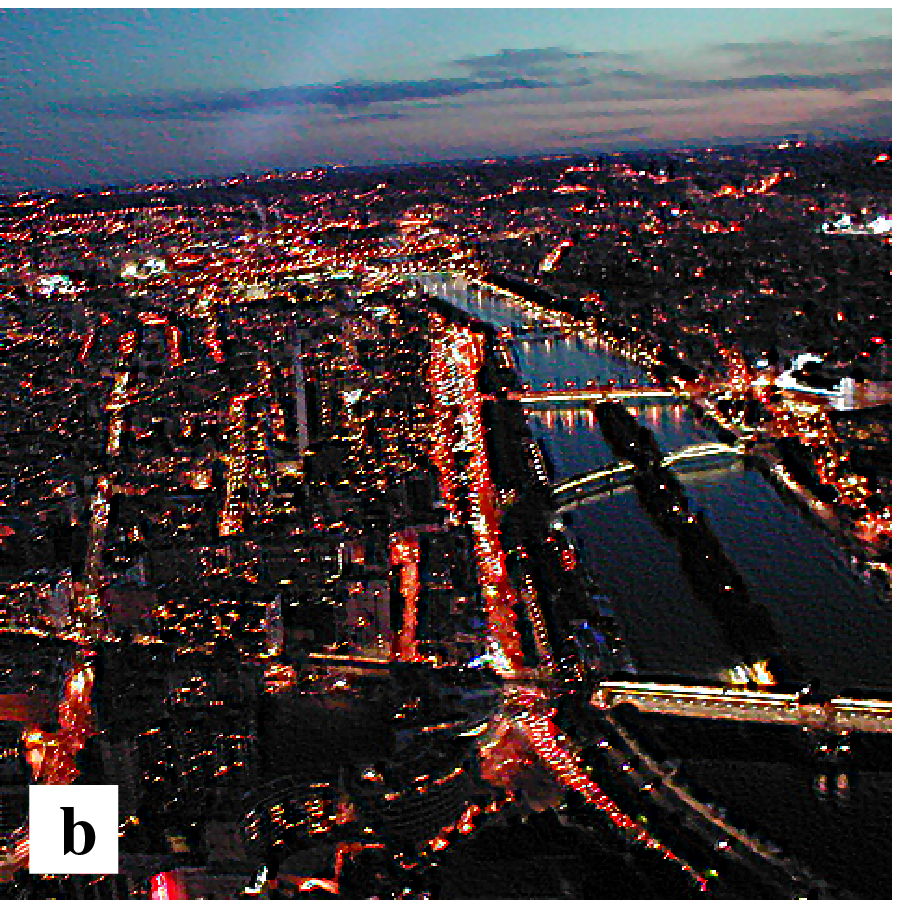}
\end{tabular}
\caption{\label{f-rwc-sivp}%
\textbf{(a)} Colour photograph (Paris from Eiffel Tower) blurred during 
acquisition, $480\times480$ pixels. Insert shows approximate PSF
used for deconvolution (same as in Figure~\ref{f-sy1-sivp}(b), 
eight times enlarged).
\textbf{(b)}~RRRL with Perona-Malik regulariser ($\lambda=13$), 
$\alpha=0.002$, $300$ iterations.
}
\end{figure}

\begin{figure}[t!]
\centering
\begin{tabular}{@{}c@{~}c@{~}c@{~}c@{}}
\includegraphics[width=0.24\textwidth]{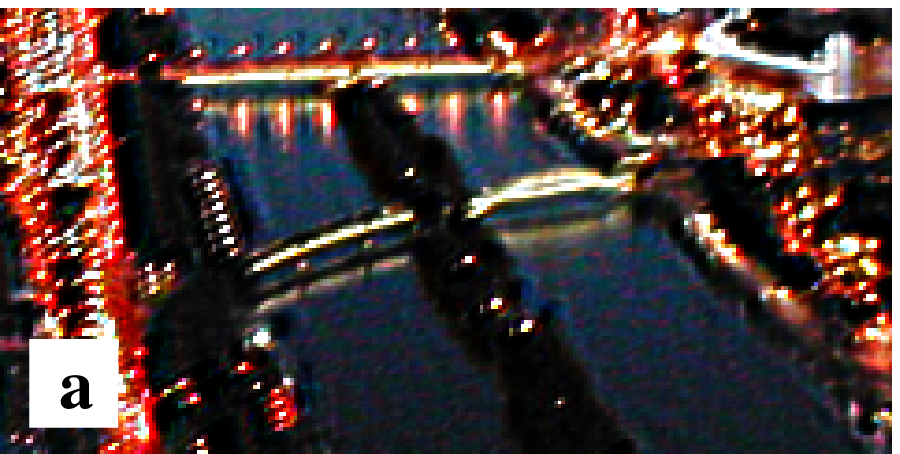}&
\includegraphics[width=0.24\textwidth]{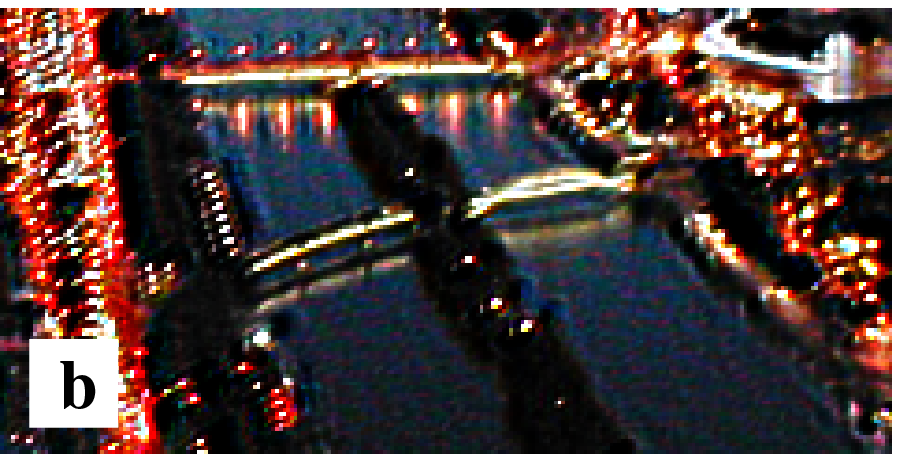}&
\includegraphics[width=0.24\textwidth]{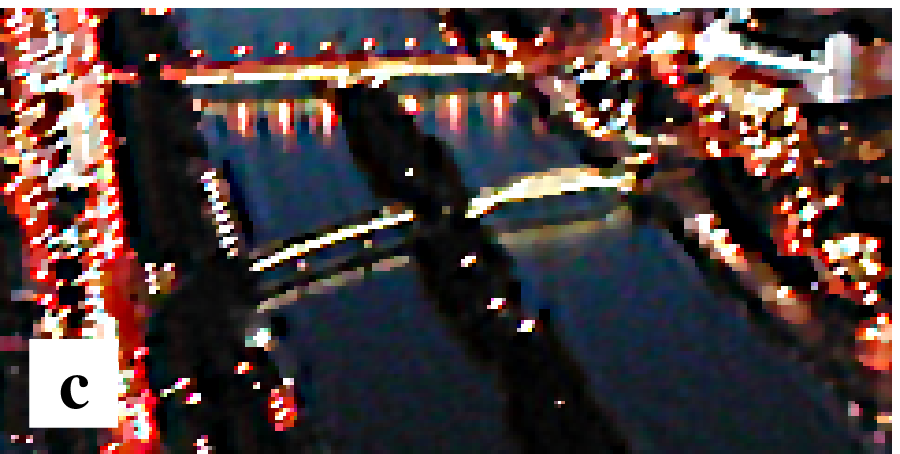}&
\includegraphics[width=0.24\textwidth]{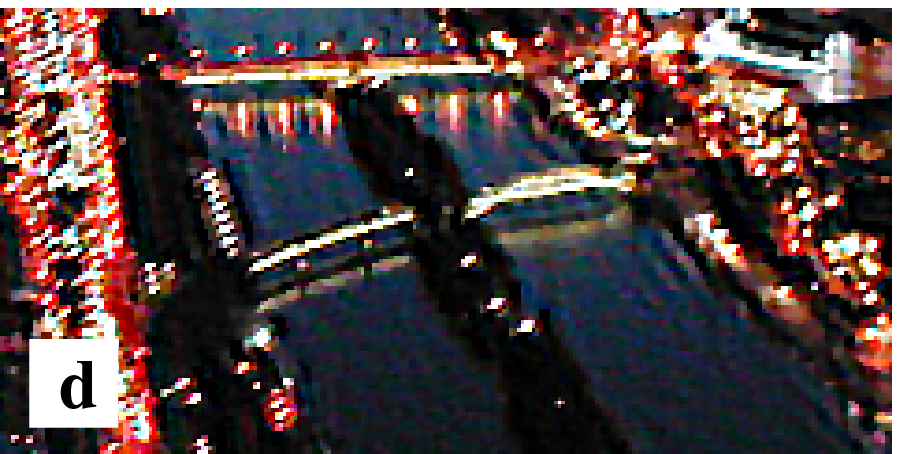}\\
\includegraphics[width=0.24\textwidth]{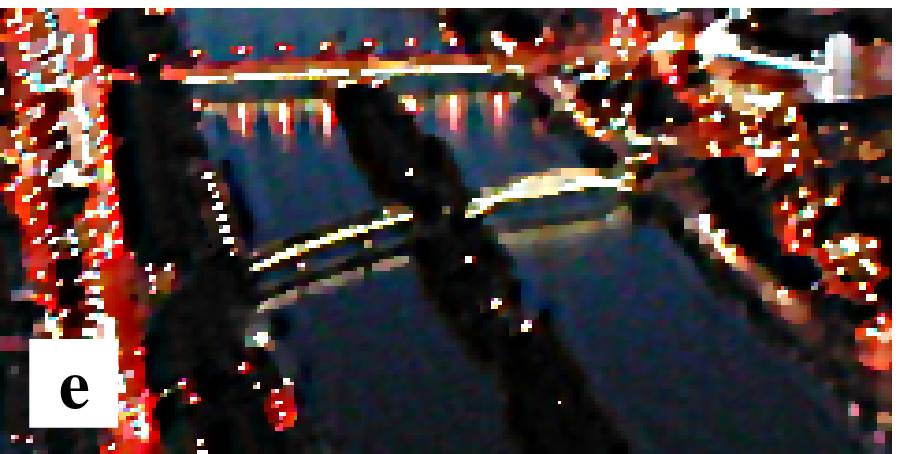}&
\includegraphics[width=0.24\textwidth]{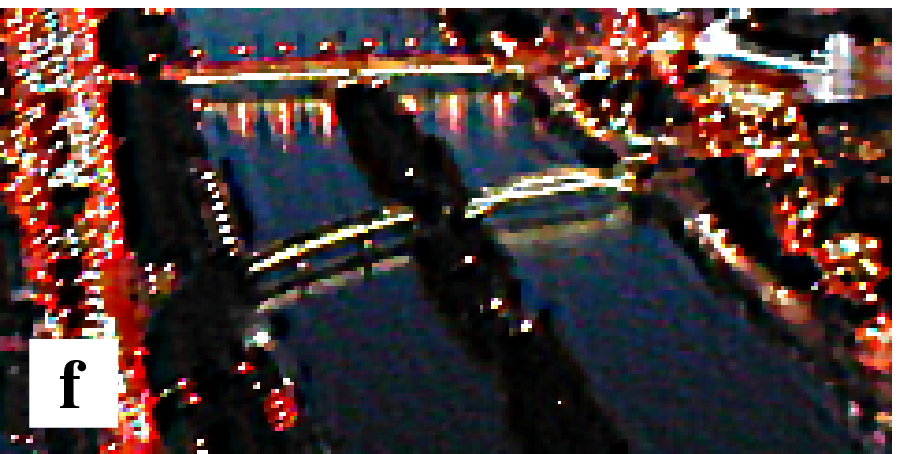}&
\includegraphics[width=0.24\textwidth]{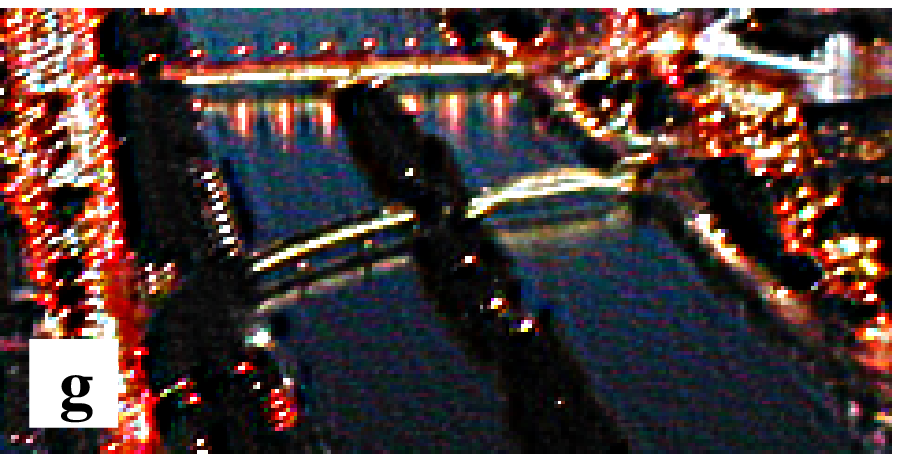}&
\includegraphics[width=0.24\textwidth]{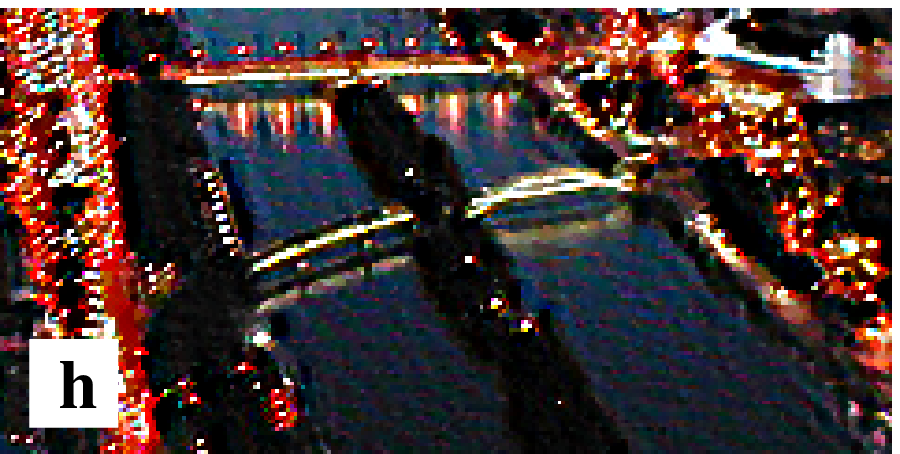}
\end{tabular}
\caption{\label{f-rwc2}%
Detail from deconvolution results for the image from 
Figure~\ref{f-rwc-sivp}(a).
\textbf{(a)}~RL, 20 iterations.
\textbf{(b)}~RL, 30 iterations.
\textbf{(c)}~Robust variational deconvolution using Perona-Malik regulariser
($\lambda=26$) without constraints,
regularisation weight $\alpha=0.06$.
\textbf{(d)}~Same but $\alpha=0.03$.
\textbf{(e)}~Robust variational deconvolution using Perona-Malik regulariser
($\lambda=26$) with positivity constraint,
regularisation weight $\alpha=0.06$.
\textbf{(f)}~Same but $\alpha=0.03$.
\textbf{(g)}~RRRL with TV regulariser, $\alpha=0.002$,
$80$ iterations.
\textbf{(h)}~RRRL with Perona-Malik regulariser
($\lambda=13$) and $\alpha=0.002$, $300$ iterations, 
see Figure~\ref{f-rwc-sivp}(b).
}
\end{figure}

\begin{table}[b!]
\caption{\label{t-rwc}%
Approximate computational expense for deconvolution results shown in 
Figures 
\ref{f-rwc-sivp} and~\ref{f-rwc2}.
Absolute computation times, which are given for rough orientation, 
were measured for single-threaded calculation on a Phenom X4 at 3.0 GHz
by a non-optimised implementation.}
\medskip
\centering
\begin{tabular}{|l|c|c|c|}
\hline
Method & Iterations & Run time & Cost factor \\
& & (s) & w.r.t.\ RL \\
\hline
Standard RL & 20 & 8.9 & 1.0 \\
Variational & 1500 & 792 & 89.0 \\
RRRL (TV) & 80 & 59.6 & 6.7 \\
RRRL (PM) & 300 & 221 & 24.8 \\
\hline
\end{tabular}
\end{table}

In order to achieve its high restoration quality,
RRRL required in both synthetic experiments significantly higher 
computational effort than standard RL. However, run times still remained 
by a factor $3\ldots5$ below those for the robust variational model from 
literature. This is caused by the favourable structure of the minimality 
condition and the fixed point iteration obtained from it. In contrast, 
minimisation of the classical variational model becomes very slow when 
getting close to the optimum, thus requiring much more iterations.

\subsection{Deconvolution of Real-World Images}

Our last experiment (Figures 
\ref{f-rwc-sivp}--\ref{f-rwc2}) is based on real-world data.
The colour photograph shown in 
Figure~\ref{f-rwc-sivp}(a)
was blurred during acquisition with an unknown point-spread function 
that is inferred approximately from the shape of a point light source. 
For restoration, we use the multi-channel versions of our methods. 

Restoration by standard RL 
achieves a decent acuity at moderate computational cost, 
see
the detail view, Figure~\ref{f-rwc2}(a). Increasing the number of iterations
quickly leads to ringing artifacts that are visible as shadows in the
vicinity of all high-contrast image structures, see Figure~\ref{f-rwc2}(b).

Variational deconvolution with a robust $L^1$ data term and Perona-Malik
regulariser allows a visible improvement in acuity over RL deconvolution
while reducing artifacts, see 
the detail view in Figure~\ref{f-rwc2}(c). 
Using the positivity-constrained gradient descent
\eqref{VGDphi} 
brings about a further significant improvement, see Figure
\ref{f-rwc2}(f). Due to the better suppression
of ringing artifacts the regularisation weight $\alpha$ could be reduced by
half here -- in contrast, unconstrained variational deconvolution with the 
same reduced $\alpha$ creates much stronger artifacts, see
Figure~\ref{f-rwc2}(d).
Imposing the constraint but retaining the larger weight $\alpha$, see
Figure~\ref{f-rwc2}(e), already improves acuity but still smoothes out
more fine details than in Figure~\ref{f-rwc2}(f).

The excellent restoration quality of variational deconvolution, however,
comes at the cost of significantly increased computation time needed in
order to approximate the steady state. 
Robust and regularised Richardson-Lucy deconvolution as shown in the last
rows of 
Figure~\ref{f-rwc2}
provides an attractive
compromise between standard RL and the variational gradient descent.
Figure~\ref{f-rwc2}(g) shows
a RRRL result with TV regulariser which can be computed fairly fast. 
With the Perona-Malik regulariser instead, see 
Figures~\ref{f-rwc-sivp}(b)
and \ref{f-rwc2}(h), 
more iterations are required in order
for the edge-enhancing properties of the regulariser to pay off, but still
the computation time is lower than with the gradient descent algorithm,
compare also Table~\ref{t-rwc}.
In terms of restoration quality, both RRRL results range between the
variational deconvolution without and with constraints. 
We remark
that the test image consisting of
large dark regions with few highlights makes the positivity constraint
particularly valuable.

\section{Conclusions}\label{sec-conc}

In this paper, 
we have investigated Richardson-Lucy deconvolution from the
variational viewpoint. Based on the observation \cite{Snyder-TIP92} that the
RL method can be understood as a fixed point iteration associated to
the minimisation of the information divergence \cite{Csiszar-AS91},
it is embedded into the framework of variational methods. 
This allows in turn to apply to it the modifications that have 
made variational deconvolution the flexible and high-quality deconvolution
tool that it is. Besides regularisation that has been proposed before
in \cite{Dey-isbi04}, we have introduced robust data terms into the model.
As a result, we have obtained a novel robust and regularised Richardson-Lucy
deconvolution method that competes in quality with state-of-the-art variational
methods, while in terms of numerical efficiency it moves considerably closer
to Richardson-Lucy deconvolution.

\bigskip
\noindent\textbf{Acknowledgement.}
Part of the work on this paper was done while the author was affiliated with
MIA group, Saarland University, Saarbr\"ucken, Germany, see \cite{Welk-tr10}.


\begin{thebibliography}{10}

\bibitem{Backes-sp09}
M.~Backes, T.~Chen, M.~D{\"u}rmuth, H.~Lensch, and M.~Welk.
\newblock Tempest in a teapot: compromising reflections revisited.
\newblock In {\em Proc.~30th IEEE Symposium on Security and Privacy}, pages
  315--327, Oakland, California, USA, 2009.

\bibitem{Bar-vlsm05}
L.~Bar, A.~Brook, N.~Sochen, and N.~Kiryati.
\newblock Color image deblurring with impulsive noise.
\newblock In N.~Paragios, O.~Faugeras, T.~Chan, and C.~Schn{\"o}rr, editors,
  {\em Variational and Level Set Methods in Computer Vision}, volume 3752 of
  {\em Lecture Notes in Computer Science}, pages 49--60. Springer, Berlin,
  2005.

\bibitem{Bar-eccv04}
L.~Bar, N.~Sochen, and N.~Kiryati.
\newblock Variational pairing of image segmentation and blind restoration.
\newblock In T.~Pajdla and J.~Matas, editors, {\em Computer Vision -- {ECCV}
  2004, Part II}, volume 3022 of {\em Lecture Notes in Computer Science}, pages
  166--177. Springer, Berlin, 2004.

\bibitem{Bar-scs05}
L.~Bar, N.~Sochen, and N.~Kiryati.
\newblock Image deblurring in the presence of salt-and-pepper noise.
\newblock In R.~Kimmel, N.~Sochen, and J.~Weickert, editors, {\em Scale Space
  and {PDE} Methods in Computer Vision}, volume 3459 of {\em Lecture Notes in
  Computer Science}, pages 107--118. Springer, Berlin, 2005.

\bibitem{Bar-ssvm07}
L.~Bar, N.~Sochen, and N.~Kiryati.
\newblock Restoration of images with piecewise space-variant blur.
\newblock In F.~Sgallari, F.~Murli, and N.~Paragios, editors, {\em Scale Space
  and Variational Methods in Computer Vision}, volume 4485 of {\em Lecture
  Notes in Computer Science}, pages 533--544. Springer, Berlin, 2007.

\bibitem{Bertero-PIEEE88}
M.~Bertero, T.~A. Poggio, and V.~Torre.
\newblock Ill-posed problems in early vision.
\newblock {\em Proceedings of the IEEE}, 76(8):869--889, Aug. 1988.

\bibitem{Bratsolis-AA01}
E.~Bratsolis and M.~Sigelle.
\newblock A spatial regularization method preserving local photometry for
  {R}ichardson-{L}ucy restoration.
\newblock {\em Astronomy and Astrophysics}, 375(3):1120--1128, 2001.

\bibitem{Chan-TIP98}
T.~F. Chan and C.~K. Wong.
\newblock Total variation blind deconvolution.
\newblock {\em IEEE Transactions on Image Processing}, 7:370--375, 1998.

\bibitem{Colton-Book92}
D.~Colton and R.~Kress.
\newblock {\em Inverse Acoustic and Electromagnetic Scattering Theory}.
\newblock Springer, Berlin, 1992.

\bibitem{Csiszar-AS91}
I.~Csisz{\'a}r.
\newblock Why least squares and maximum entropy? an axiomatic approach to
  inference for linear inverse problems.
\newblock {\em Annals of Statistics}, 19(4):2032--3066, 1991.

\bibitem{Dey-isbi04}
N.~Dey, L.~Blanc-F{\'e}raud, C.~Zimmer, Z.~Kam, J.-C. Olivo-Marin, and
  J.~Zerubia.
\newblock A deconvolution method for confocal microscopy with total variation
  regularization.
\newblock In {\em Proc. IEEE International Symposium on Biomedical Imaging
  (ISBI)}, April 2004.

\bibitem{Dey-MRT06}
N.~Dey, L.~Blanc-Feraud, C.~Zimmer, P.~Roux, Z.~Kam, J.-C. Olivo-Marin, and
  J.~Zerubia.
\newblock {R}ichardson-{L}ucy algorithm with total variation regularization for
  {3D} confocal microscope deconvolution.
\newblock {\em Microscopy Research and Technique}, 69:260--266, 2006.

\bibitem{Elhayek-dagm11}
A.~Elhayek, M.~Welk, and J.~Weickert.
\newblock Simultaneous interpolation and deconvolution model for the 3-{D}
  reconstruction of cell images.
\newblock In R.~Mester and M.~Felsberg, editors, {\em Pattern Recognition},
  volume 6835 of {\em Lecture Notes in Computer Science}, pages 316--325.
  Springer, Berlin, 2011.

\bibitem{Geman-PAMI84}
S.~Geman and D.~Geman.
\newblock Stochastic relaxation, {G}ibbs distributions, and the {B}ayesian
  restoration of images.
\newblock {\em IEEE Transactions on Pattern Analysis and Machine Intelligence},
  6:721--741, 1984.

\bibitem{Gerig-TMI92}
G.~Gerig, O.~K\"ubler, R.~Kikinis, and F.~A. Jolesz.
\newblock Nonlinear anisotropic filtering of {MRI} data.
\newblock {\em IEEE Transactions on Medical Imaging}, 11:221--232, 1992.

\bibitem{Hirsch-iccv11}
M.~Hirsch, C.~J. Schuler, S.~Harmeling, and B.~Sch{\"o}lkopf.
\newblock Fast removal of non-uniform camera shake.
\newblock In {\em Proc. IEEE International Conference on Computer Vision},
  pages 463--470, Barcelona, 2011.

\bibitem{Huber-book81}
P.~J. Huber.
\newblock {\em Robust Statistics}.
\newblock Wiley, New York, 1981.

\bibitem{Jung-TIP11}
M.~Jung, X.~Bresson, T.~F. Chan, and L.~A. Vese.
\newblock Nonlocal {M}umford--{S}hah regularizers for color image restoration.
\newblock {\em IEEE Transactions on Image Processing}, 20(6):1583--1598, 2011.

\bibitem{Jung-JCAM13}
M.~Jung, A.~Marquina, and L.~A. Vese.
\newblock Variational multiframe restoration of images degraded by noisy
  (stochastic) blur kernels.
\newblock {\em Journal of Computational and Applied Mathematics},
  240(1):123--134, 2013.

\bibitem{Jung-ssvm09}
M.~Jung and L.~A. Vese.
\newblock Nonlocal variational image deblurring models in the presence of
  {G}aussian or impulse noise.
\newblock In X.-C. Tai, K.~M{\o}rken, M.~Lysaker, and K.-A. Lie, editors, {\em
  Scale-Space and Variational Methods in Computer Vision}, volume 5567 of {\em
  Lecture Notes in Computer Science}, pages 402--413. Springer, Berlin, 2009.

\bibitem{Krishnan-nips09}
D.~Krishnan and R.~Fergus.
\newblock Fast image deconvolution using hyper-laplacian priors.
\newblock In {\em Advances in Neural Information Processing Systems}, pages
  1033--1041, 2009.

\bibitem{Lucy-AJ74}
L.~B. Lucy.
\newblock An iterative technique for the rectification of observed
  distributions.
\newblock {\em The Astronomical Journal}, 79(6):745--754, June 1974.

\bibitem{Nagy-spie00}
J.~G. Nagy and Z.~Strako{\v s}.
\newblock Enforcing nonnegativity in image reconstruction algorithms.
\newblock In D.~C. Wilson, H.~D. Tagare, F.~L. Bookstein, F.~J. Preteux, and
  E.~R. Dougherty, editors, {\em Advanced Signal Processing Algorithms,
  Architectures, and Implementations}, volume 4121 of {\em Proceedings of
  {SPIE}}, pages 182--190. SPIE Press, Bellingham, 2000.

\bibitem{Osher-icip94}
S.~Osher and L.~Rudin.
\newblock Total variation based image restoration with free local constraints.
\newblock In {\em Proc.~1994 IEEE International Conference on Image
  Processing}, pages 31--35, Austin, Texas, 1994.

\bibitem{Perona-PAMI90}
P.~Perona and J.~Malik.
\newblock Scale space and edge detection using anisotropic diffusion.
\newblock {\em IEEE Transactions on Pattern Analysis and Machine Intelligence},
  12:629--639, 1990.

\bibitem{Persch-MST13x}
N.~Persch, A.~Elhayek, M.~Welk, A.~Bruhn, S.~Grewenig, K.~B{\"o}se,
  A.~Kraegeloh, and J.~Weickert.
\newblock Enhancing 3-{D} cell structures in confocal and {STED} microscopy: a
  joint model for interpolation, deblurring and anisotropic smoothing.
\newblock {\em Measurement Science and Technology}, in press, 2013.

\bibitem{Richardson-JOSA72}
W.~H. Richardson.
\newblock Bayesian-based iterative method of image restoration.
\newblock {\em Journal of the Optical Society of America}, 62(1):55--59, 1972.

\bibitem{Rudin-PHYSD92}
L.~I. Rudin, S.~Osher, and E.~Fatemi.
\newblock Nonlinear total variation based noise removal algorithms.
\newblock {\em Physica D}, 60:259--268, 1992.

\bibitem{Sawatzky-aip10}
A.~Sawatzky and M.~Burger.
\newblock Edge-preserving regularization for the deconvolution of biological
  images in nanoscopy.
\newblock In G.~Psihoyios, T.~Simos, and C.~Tsitouras, editors, {\em Proc.~8th
  International Conference of Numerical Analysis and Applied Mathematics},
  volume 1281 of {\em Conference Proceedings}, pages 1983--1986. AIP, September
  2010.

\bibitem{Snyder-TIP92}
D.~Snyder, T.~J. Schulz, and J.~A. O'Sullivan.
\newblock Deblurring subject to nonnegativity constraints.
\newblock {\em IEEE Transactions on Image Processing}, 40(5):1143--1150, 1992.

\bibitem{Tikhonov-SMD63}
A.~N. Tikhonov.
\newblock Solution of incorrectly formulated problems and the regularization
  method.
\newblock {\em Soviet Mathematics Doklady}, 4:1035--1038, 1963.

\bibitem{vanCittert-ZPhys33}
P.~H. {van Cittert}.
\newblock Zum {E}influ{\ss} der {S}paltbreite auf die
  {I}ntensit{\"a}tsverteilung in {S}pektrallinien. {II.}
\newblock {\em Zeitschrift f{\"u}r Physik}, 65:298--308, 1933.

\bibitem{Vogel-TIP98}
C.~R. Vogel and M.~E. Oman.
\newblock Fast, robust total variation-based reconstruction of noisy, blurred
  images.
\newblock {\em IEEE Transactions on Image Processing}, 7:813--824, 1998.

\bibitem{Wang-SIIMS08}
Y.~Wang, J.~Yang, W.~Yin, and Y.~Zhang.
\newblock A new alternating minimization algorithm for total variation image
  reconstruction.
\newblock {\em SIAM Journal on Imaging Sciences}, 1(3):248--272, 2008.

\bibitem{Welk-tr10}
M.~Welk.
\newblock Robust variational approaches to positivity-constrained image
  deconvolution.
\newblock Technical Report 261, Department of Mathematics, Saarland University,
  Saarbr{\"u}cken, Germany, Mar. 2010.

\bibitem{Welk-aapr13}
M.~Welk and M.~Erler.
\newblock Algorithmic optimisations for iterative deconvolution methods.
\newblock In J.~Piater and A.~Rodr{\'\i}guez-S{\'a}nchez, editors, {\em
  Proceedings of the 37th {A}nnual {W}orkshop of the {A}ustrian {A}ssociation
  for {P}attern {R}ecognition ({{\"O}AGM/AAPR}), 2013}. arXiv:1304.7211
  [cs.CV], 2013.

\bibitem{Welk-ibpria07}
M.~Welk and J.~G. Nagy.
\newblock Variational deconvolution of multi-channel images with inequality
  constraints.
\newblock In J.~Mart{\'\i}, J.~M. Bened{\'\i}, A.~M. Mendon{\c c}a, and
  J.~Serrat, editors, {\em Pattern Recognition and Image Analysis}, volume 4477
  of {\em Lecture Notes in Computer Science}, pages 386--393. Springer, Berlin,
  2007.

\bibitem{Welk-SIVP13}
M.~Welk, P.~Raudaschl, T.~Schwarzbauer, M.~Erler, and M.~L{\"a}uter.
\newblock Fast and robust linear motion deblurring.
\newblock {\em Signal, Image and Video Processing}, 2013, 
DOI 10.1007/s11760-013-0563-x (online first).

\bibitem{Welk-TBW-scs05}
M.~Welk, D.~Theis, T.~Brox, and J.~Weickert.
\newblock {PDE}-based deconvolution with forward-backward diffusivities and
  diffusion tensors.
\newblock In R.~Kimmel, N.~Sochen, and J.~Weickert, editors, {\em Scale Space
  and {PDE} Methods in Computer Vision}, volume 3459 of {\em Lecture Notes in
  Computer Science}, pages 585--597. Springer, Berlin, 2005.

\bibitem{Welk-dagm05}
M.~Welk, D.~Theis, and J.~Weickert.
\newblock Variational deblurring of images with uncertain and spatially variant
  blurs.
\newblock In W.~Kropatsch, R.~Sablatnig, and A.~Hanbury, editors, {\em Pattern
  Recognition}, volume 3663 of {\em Lecture Notes in Computer Science}, pages
  485--492. Springer, Berlin, 2005.

\bibitem{Whittaker-PEMS23}
E.~T. Whittaker.
\newblock On a new method of graduation.
\newblock {\em Proceedings of the Edinburgh Mathematical Society}, 41:63--75,
  1923.

\bibitem{Wiener-Book49}
N.~Wiener.
\newblock {\em Extrapolation, Interpolation and Smoothing of Stationary Time
  Series with Engineering Applications}.
\newblock {MIT} Press, Cambridge, {MA}, 1949.

\bibitem{You-icip96}
Y.-L. You and M.~Kaveh.
\newblock Anisotropic blind image restoration.
\newblock In {\em Proc.~1996 IEEE International Conference on Image
  Processing}, volume~2, pages 461--464, Lausanne, Switzerland, Sept. 1996.

\bibitem{Zervakis-TIP95}
M.~E. Zervakis, A.~K. Katsaggelos, and T.~M. Kwon.
\newblock A class of robust entropic functionals for image restoration.
\newblock {\em IEEE Transactions on Image Processing}, 4(6):752--773, June
  1995.

\end{thebibliography}
\end{document}